\PassOptionsToPackage{table}{xcolor}
\documentclass[sigconf, screen]{acmart}
\AtBeginDocument{%
  }

\setcopyright{acmlicensed}
\copyrightyear{2018}
\acmYear{2025}
\acmDOI{XXXXXXX.XXXXXXX}
\acmConference[Conference acronym 'XX]{Make sure to enter the correct
  conference title from your rights confirmation email}{June 03--05,
  2018}{Woodstock, NY}
\acmISBN{978-1-4503-XXXX-X/2018/06}

\usepackage{bbding}
\usepackage{multirow}
\usepackage{pgfplots}
\usepackage{pgfplotstable}
\usepgfplotslibrary{polar}
\usetikzlibrary{positioning}
\usepackage{graphicx}
\usepackage{subcaption}
\usepackage[table]{xcolor}
\usepackage{colortbl}
\usepackage{tikz}
\usepackage{longtable,multirow,array,booktabs}
\usepackage{enumitem}
\usepackage{tcolorbox}

\makeatletter
\renewcommand\@makefntext[1]{%
    \noindent#1}
\makeatother
\begin{document}
\definecolor{darkgreen}{rgb}{0.0, 0.5, 0.0}
\definecolor{darkyellow}{rgb}{0.6, 0.5, 0.0}
\definecolor{LightCyan}{rgb}{0.80, 1, 1}
\definecolor{LightPink}{rgb}{1, 0.89, 0.88}

\newcommand{\highlight}[2]{\tikz[baseline]{\node[fill=#1, anchor=base, minimum width=2em, inner sep=1pt, outer sep=0.5pt, rounded corners] {#2};}}

\title{GeoSense: Evaluating Identification and Application of Geometric Principles in Multimodal Reasoning}

\author{
  Liangyu Xu\textsuperscript{$\ast$}, 
  Yingxiu Zhao\textsuperscript{$\ast$}, 
  Jingyun Wang, 
  Yingyao Wang \\ 
  \and 
  Pi Bu, 
  Chen Wang, 
  Mingliang Zhang, 
  Jihao Gu \\ 
  \and 
  Xiang Li, 
  Xiaoyong Zhu, 
  Jun Song\textsuperscript{$\dag$}, 
  Bo Zheng
}

\affiliation{%
  \institution{Taobao \& Tmall Group of Alibaba} 
  \city{Beijing} \country{China}
}

\thanks{$\ast$ Equal contributions}
\thanks{$\dag$ Corresponding author: Jun Song. Email: jsong.sj@taobao.com}

\begin{abstract}
Geometry problem-solving (GPS), a challenging task requiring both visual comprehension and symbolic reasoning, effectively measures the reasoning capabilities of multimodal large language models (MLLMs).
Humans exhibit strong reasoning ability in this task through accurate identification and adaptive application of geometric principles within visual contexts. 
However, existing benchmarks fail to jointly assess both dimensions of the human-like geometric reasoning mechanism in MLLMs, remaining a critical gap in assessing their ability to tackle GPS. To this end, we introduce \textbf{GeoSense}, the first comprehensive bilingual benchmark designed to systematically evaluate the geometric reasoning abilities of MLLMs through the lens of geometric principles. GeoSense features a five-level hierarchical framework of geometric principles spanning plane and solid geometry, an intricately annotated dataset of 1,789 problems, and an innovative evaluation strategy. Through extensive experiments on GeoSense with various open-source and closed-source MLLMs, we observe that Gemini-2.0-pro-flash performs best, achieving an overall score of $65.3$.
Our in-depth analysis reveals that the identification and application of geometric principles remain a bottleneck for leading MLLMs, jointly hindering their reasoning abilities. These findings underscore GeoSense's potential to guide future advancements in MLLMs' geometric reasoning capabilities, paving the way for more robust and human-like reasoning in artificial intelligence.
We will open-source the code and dataset within a month.

\end{abstract}


\begin{teaserfigure}
   \centering
  \begin{subfigure}[b]{0.33\linewidth}
    \centering
    \includegraphics[width=\linewidth]{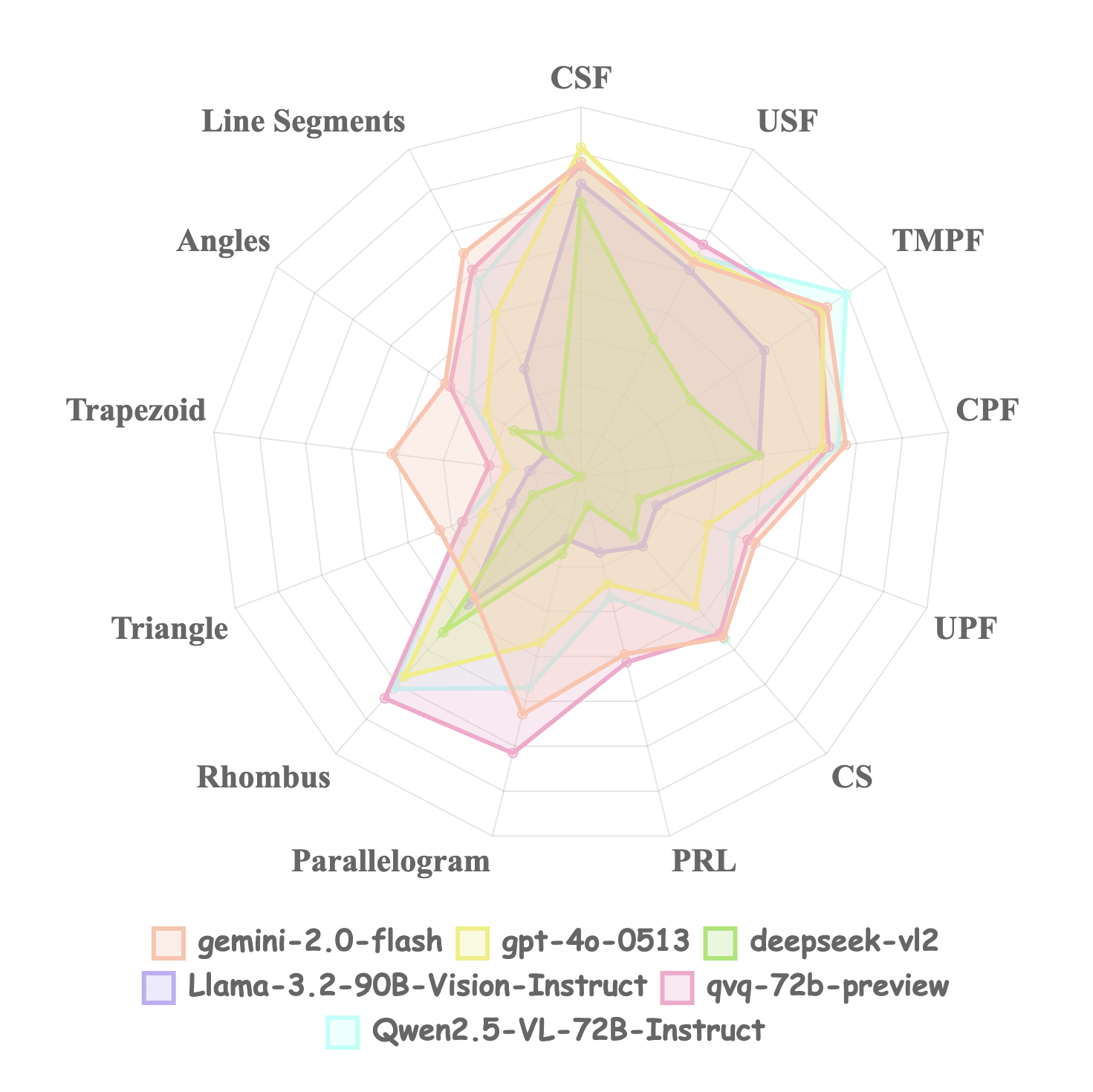}
    \caption{GPI Performance.}
    \Description{A woman and a girl in white dresses sit in an open car.}
    \label{fig:radargpi}
  \end{subfigure}
  \begin{subfigure}[b]{0.33\linewidth}
    \centering
    \includegraphics[width=\linewidth]{figs/GPA.pdf}
    \caption{GPA Performance.}
    \Description{A woman and a girl in white dresses sit in an open car.}
    \label{fig:radargpa}
  \end{subfigure}
  \begin{subfigure}[b]{0.33\linewidth}
    \centering
    \includegraphics[width=\linewidth]{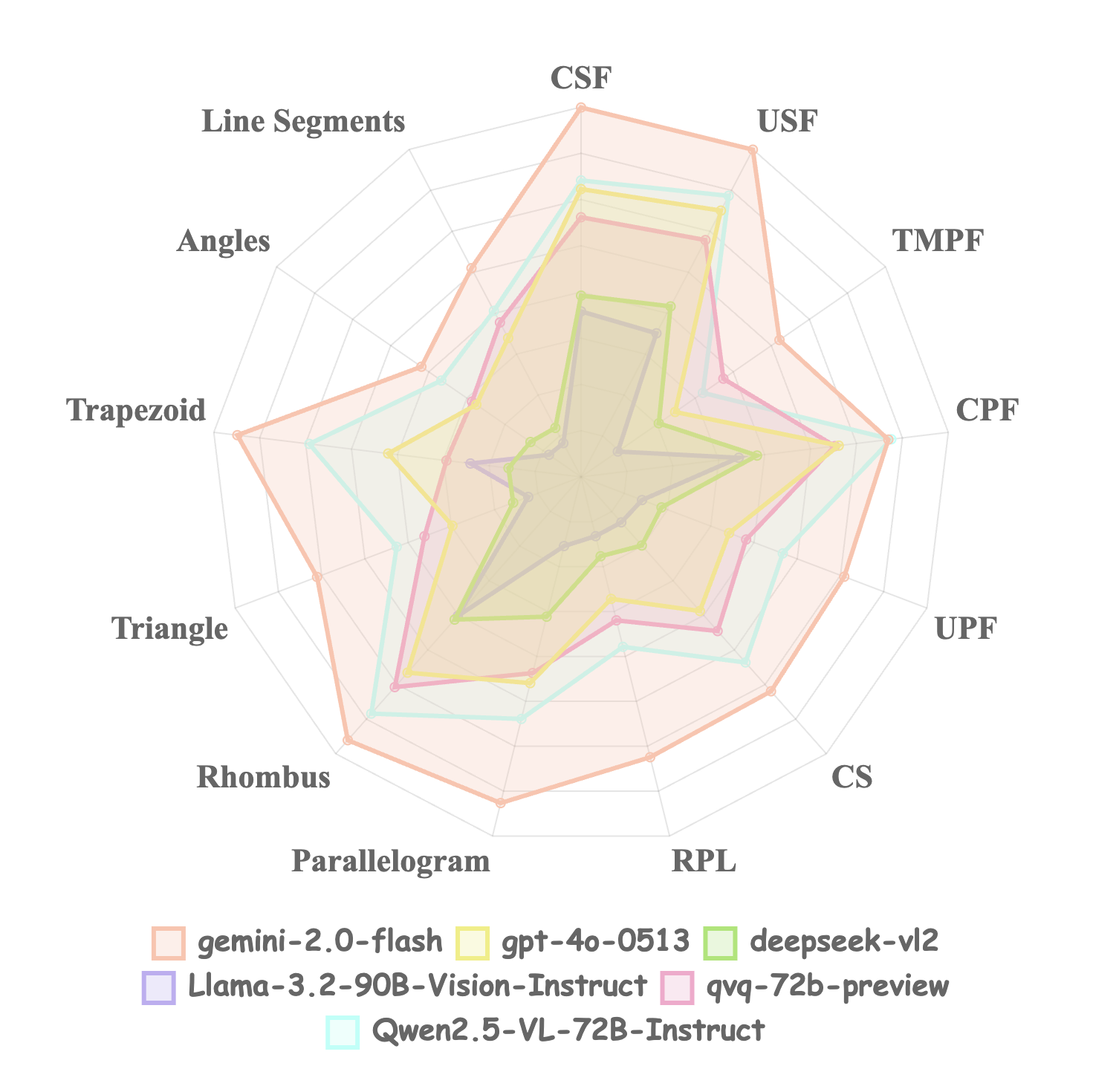}
    \caption{ACC Performance.}
    \Description{A woman and a girl in white dresses sit in an open car.}
    \label{fig:radaracc}
  \end{subfigure}
  \caption{The performance of MLLMs in different subjects across (a) GPI, (b) GPA, and (c) ACC. }
  \label{fig}
\end{teaserfigure}

\settopmatter{printacmref=false} 
\renewcommand\footnotetextcopyrightpermission[1]{} 
\pagestyle{plain} 
\setcopyright{none}

\maketitle

\section{Introduction}

\begin{figure}[htbp]
  \centering
  \includegraphics[width=\linewidth]{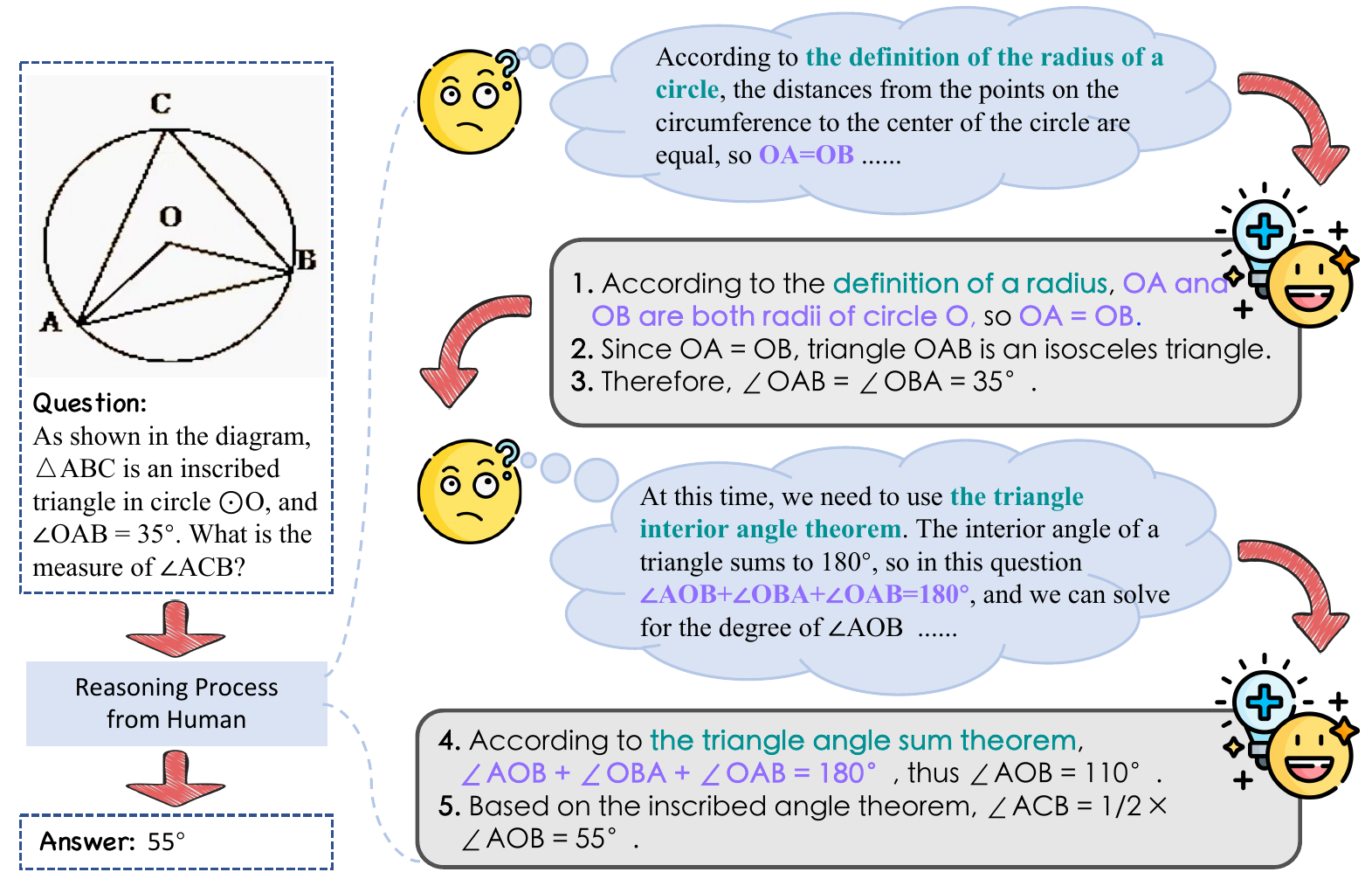}
  \caption{Humans solve geometric problems by first identifying the relevant geometric principles and then applying them to derive solutions.}
  \label{fig:human}
  \vspace{-0.5cm}
\end{figure}

\underline{G}eometry \underline{P}roblem-\underline{S}olving (GPS) involves understanding spatial relationships and employing symbolic reasoning to arrive at logical solutions within complex visual contexts\cite{lu2021intergpsinterpretablegeometryproblem,chen2022geoqageometricquestionanswering,kazemi2023geomversesystematicevaluationlarge, zhang2024mathversedoesmultimodalllm, qiao2024wemathdoeslargemultimodal}.  
When faced with geometry problems, humans exhibit exceptional skills by correctly identifying geometric principles\footnote{Geometric principles include foundational geometric concepts such as definitions, theorems, and formulas. See details in Section~\ref{sec:2.0}.} and then adaptively applying them to derive solutions, as illustrated in Fig.~\ref{fig:human}. 

\begin{figure}[t]
  \centering
  \includegraphics[width=\linewidth]{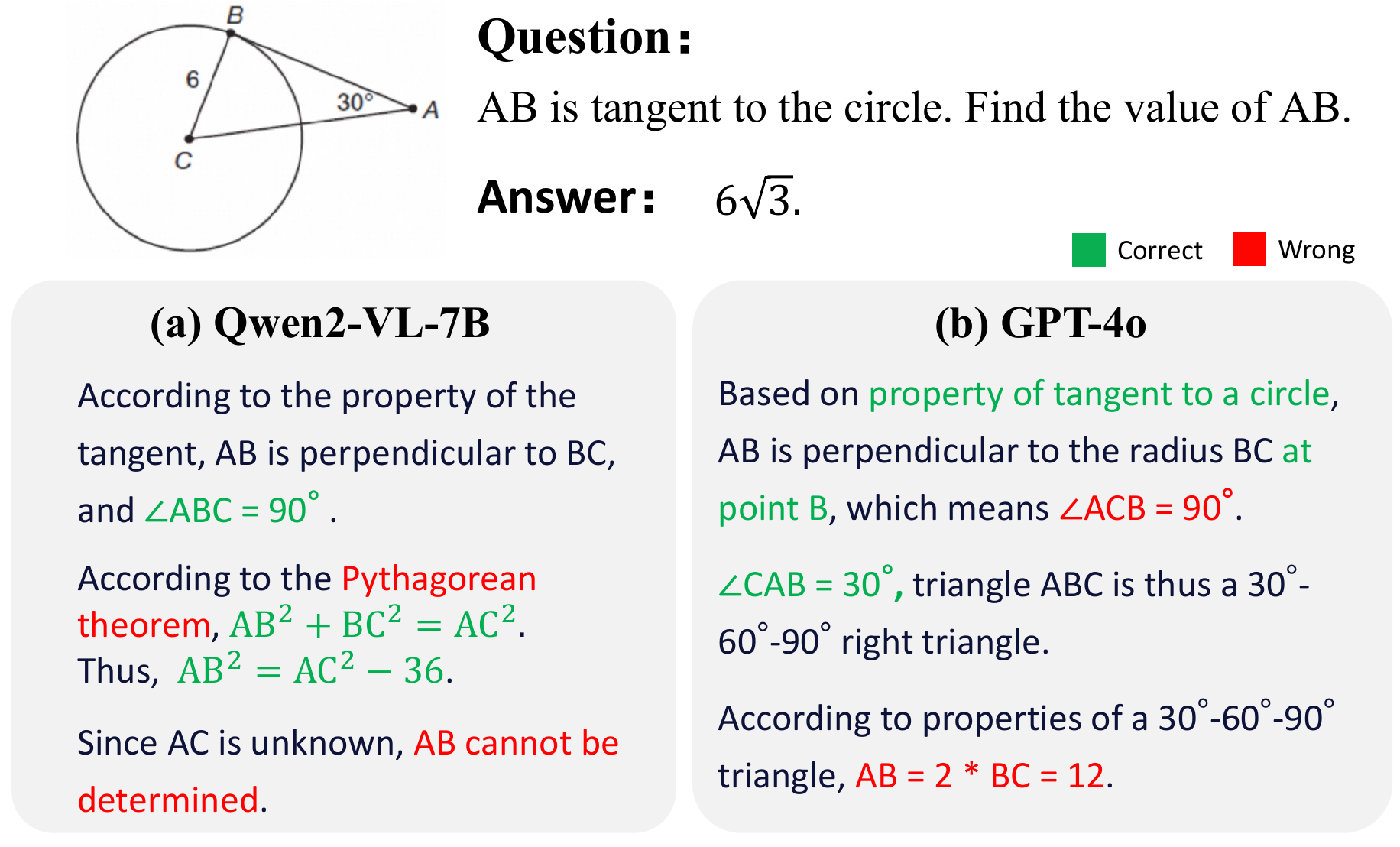}
  \vspace{-0.5cm}
  \caption{MLLMs encounter failures in GPS: Qwen2-VL-7B fails to identify the correct principle and GPT-4o struggles to apply principles to solve questions.}
  \vspace{-0.2cm}
  \label{fig:case}
\end{figure}

With the rise of multimodal large language models (MLLMs)~\cite{geminiteam2024geminifamilyhighlycapable,openai2023gpt4v,bai2023qwenvlversatilevisionlanguagemodel,wang2024qwen2vlenhancingvisionlanguagemodels,liu2023llava}, GPS emerges as a crucial testbed for evaluating the reasoning capabilities of MLLMs. 
Numerous benchmarks are developed to assess MLLMs' performance on GPS, with most prior studies~\cite{seo-etal-2015-solving, lu2021intergpsinterpretablegeometryproblem,chen2022geoqageometricquestionanswering, lu2024mathvistaevaluatingmathematicalreasoning} primarily focusing on the correctness of final answers. 
Recent researches shift attention to the reasoning process of MLLMs in GPS~\cite{zhang2024mathversedoesmultimodalllm, qiao2024wemathdoeslargemultimodal, chen2025mathflowenhancingperceptualflow}. 
Despite progress, challenges persist. 
For instance, studies like MathVerse~\cite{zhang2024mathversedoesmultimodalllm} and MathFlow~\cite{ chen2025mathflowenhancingperceptualflow} point out that MLLMs' imprecise perception and interpretation of geometric diagrams can impede subsequent reasoning.
Nonetheless, through extensive experiments and analysis, we further reveal that even when MLLMs demonstrate accurate visual perception and logical reasoning steps, their lack of essential geometric principles often leads to failure, as shown in Fig.~\ref{fig:case} (a).
Moreover, while WeMath~\cite{qiao2024wemathdoeslargemultimodal} acknowledges the importance of underlying principles, it falls short of evaluating whether MLLMs correctly apply them within the visual context.
For example, even with correct retrieval of geometric principles, incorrect application can still result in reasoning failures, as shown in Fig.~\ref{fig:case} (b).
These underscore a critical gap in current research:
There is a lack of comprehensive evaluation framework to assess simultaneously both the accurate identification of geometric principles and their proper contextual application within complicated visual scenarios.

In light of these limitations, we propose two key questions for evaluating MLLMs in GPS tasks:

\noindent\textbf{Q1: }\textit{ Could MLLMs \textbf{accurately identify} the required geometric principles when solving geometric problems? }

\noindent\textbf{Q2: } \textit{Are MLLMs capable of \textbf{adaptively applying} geometric principles to visual geometric diagrams?  }

To address these questions, we propose \textbf{GeoSense}, the first comprehensive bilingual benchmark designed to systematically assess the reasoning abilities of MLLMs through the perspective of geometric principles. 
GeoSense features a hierarchical system of geometric principles, an intricately structured and finely annotated dataset, and an innovative evaluation strategy. 
It supports both English and Chinese, bridging human intuition and MLLMs' reasoning.
Specifically, we first organize geometric principles into a multi-level framework consisting of 148 unique principles, spanning from plane geometry to solid geometry. 
This framework encompasses graphical computation, conceptual understanding, and detailed definitions, theorems, and formulas. 
Next, we curate a dataset of 1,789 geometric math problems sourced from existing benchmarks and the IXL website\footnote{https://www.ixl.com/math/geometry}. 
To ensure precise annotation, we employ 23 expert annotators and develop a semi-automated annotation pipeline, ultimately generating 5,556 geometric principles and their aligned application within geometric diagrams.
Furthermore, we propose a novel evaluation strategy with two key evaluation metrics: \textsl{Geometry Principle Identification (GPI)}, which assesses MLLMs' ability to identify the most relevant geometric principles for a given problem; and \textsl{Geometry Principle Application (GPA)}, which evaluates their skills in aligning and applying these principles to specific elements in geometric diagrams. 
Additionally, we use answer accuracy (ACC) as a holistic measure of a model's overall performance. 
Table~\ref{tab:compare} compares GeoSense with other GPS-related reasoning benchmarks.

Extensive experiments on GeoSense are performed with various popular open-source and closed-source MLLMs, yielding several key insights.
As shown in Fig. \ref{fig}, while MLLMs excel in computation tasks, they struggle with understanding tasks, which can be proven by lower GPI scores in plane geometry.
Notably, though Gemini-2.0-Pro-Flash outperforms others by final answer accuracy, its GPA scores, representing the capability of adaptively applying geometric principles within visual contexts, are still limited.
Such a result further verifies that our proposed GeoSense points out new pathways for MLLMs' future advancements in GPS.





In summary, our contributions are outlined as follows:
\begin{itemize}[leftmargin=*]
 \item{We develop GeoSense, the first comprehensive bilingual benchmark that systematically evaluates the reasoning abilities of MLLMs rooted in geometric principles.}
 \item{We establish a holistic framework for geometric principles, providing a structured hierarchy for geometric reasoning.}
 \item{We design an innovative evaluation strategy to thoroughly assess the MLLMs' ability to identify and effectively apply them within visual diagrams.}
 \item{Our extensive experiments and analysis yield valuable insights for enhancing MLLMs reasoning abilities for solving geometry problems.}
\end{itemize}






\begin{table*}[htbp]
\begin{tabular}{l|cccccccccc}
\hline
\textbf{BenchMark}   & \textbf{Size} & \textbf{Language} & \textbf{Category} &  \textbf{Level} & \textbf{CoT-E.} & \textbf{P.I.} & \textbf{P.A.} & \textbf{Metric} \\ \hline
\textbf{GeoS \cite{seo-etal-2015-solving}} & 186 & EN                & -  & SAT & \textcolor{red}{$\times$}           & \textcolor{red}{$\times$}            & \textcolor{red}{$\times$}  & Rule-based      \\
\textbf{Geometry3K \cite{lu2021intergpsinterpretablegeometryproblem}} & 601   & EN  & 4  & Middle School & \textcolor{red}{$\times$}           & \textcolor{red}{$\times$}             & \textcolor{green}{$\checkmark$}    & Rule-based     \\
\textbf{GeoQA \cite{chen2022geoqageometricquestionanswering}}   & 754 & EN\&CH   &  3  & Middle School  & \textcolor{red}{$\times$}          & \textcolor{green}{$\checkmark$}            & \textcolor{red}{$\times$}  & Rule-based
 \\
\textbf{GeoQA+ \cite{cao-xiao-2022-augmented}}       & 755             & EN\&CH     &   3        & Middle School                & \textcolor{red}{$\times$}          & \textcolor{green}{$\checkmark$}            & \textcolor{red}{$\times$}     & Rule-based   \\
\textbf{PGPS9K \cite{zhang2023multimodalneuralgeometricsolver}} & 1,000             & EN   &      30      &     Middle School  & \textcolor{red}{$\times$}           & \textcolor{red}{$\times$}             & \textcolor{green}{$\checkmark$}  & Rule-based       \\
\textbf{GeoEval \cite{zhang2024geoevalbenchmarkevaluatingllms}}     & 2,000 & EN\&CH  & 7  & Middle\&High School  & \textcolor{red}{$\times$}   & \textcolor{green}{$\checkmark$} & \textcolor{red}{$\times$} & Rule-based   \\
\textbf{GeomVerse \cite{kazemi2023geomversesystematicevaluationlarge}}  & 2,000            & EN       &  5      & Synthetic         & \textcolor{red}{$\times$}           & \textcolor{red}{$\times$}             & \textcolor{green}{$\checkmark$} & Rule-based \\
\textbf{MathVision \cite{zhang2024mathversedoesmultimodalllm}}  & 2,612            & EN                & 12      & Competitions          & \textcolor{green}{$\checkmark$}           & \textcolor{red}{$\times$}           & \textcolor{red}{$\times$}         & LLM-as-a-judge  \\
\textbf{MathVista \cite{lu2024mathvistaevaluatingmathematicalreasoning}}  & 5,487            & EN                & 19 & SAT\&Middle School               & \textcolor{red}{$\times$}           & \textcolor{red}{$\times$}            & \textcolor{red}{$\times$}            &   LLM-as-a-judge     \\
\textbf{MathVerse \cite{zhang2024mathversedoesmultimodalllm}}  & 2,612            & EN                & 12          &    High School   & \textcolor{green}{$\checkmark$}           & \textcolor{red}{$\times$}           & \textcolor{red}{$\times$}     & LLM-as-a-judge      \\
\textbf{WeMath \cite{qiao2024wemathdoeslargemultimodal}}    & 1,674            & EN                & 67     &       Middle\&High School    & \textcolor{red}{$\times$}           & \textcolor{green}{$\checkmark$}            & \textcolor{red}{$\times$}    & Rule-based               \\

\hline

\textbf{GeoSense} & 1,789  & EN\&CH  & 148 & Middle\&High School & \textcolor{green}{$\checkmark$}   & \textcolor{green}{$\checkmark$}             & \textcolor{green}{$\checkmark$}  & LLM-as-a-judge\\
\hline
\end{tabular}
\caption{Comparisons between our GeoSense and other GPS-related benchmarks. CoT-E., P.I. and P.A. are short for CoT evaluation, principle identification, and principle application, respectively.}
\label{tab:compare}
\vspace{-0.5cm}
\end{table*}

\section{The GeoSense Dataset}
We first give a clear definition of geometric principles in Section~\ref{sec:2.0}.
Then, in Section~\ref{sec:2.1}, we provide an overview of the dataset, including its composition, categorization, and detailed statistical information.
We also construct a fine-grained hierarchical framework of geometric principles.
Furthermore, in Section~\ref{sec:2.2}, we introduce a pipeline for the annotation of essential geometric principles and their application in visual contexts for our collected data.

\subsection{Definition of Geometric Principles}\label{sec:2.0}
\texttt{"Euclid's elements\footnote{https://en.wikipedia.org/wiki/Euclid\%27s\_Elements}"} provides a foundational classification of geometric propositions, including definitions, theorems, postulates, and axioms. 
Drawing inspiration from Euclid's systematic approach, we focus on the most commonly utilized categories and define geometric principles as fundamental concepts in geometry. These principles encapsulate definitions, theorems, and formulas, with the latter specifically representing mathematical expressions for calculations, such as 
the calculation of length, area, and volume.

\subsection{Data Collection and Statistics} \label{sec:2.1}
\begin{table}[t]
\small
  \centering
  \resizebox{0.87\linewidth}{!}{
    \begin{tabular}{lr}
    \toprule
    \textbf{Statistic} & \textbf{Number} \\
    \midrule
    Total questions & 1,789 \\
    - Multiple-choice questions & 231 (12.9\%) \\
    - Free-form question &  1,558 (87.1\%)\\
    - Newly collected questions & 673 (37.6 \%) \\
    - Existing-dataset questions & 1,116 (62.4 \%) \\
    \midrule
    Annotated geometric principles & 5,556 \\
    - Definitions & 3,235 (58.2\%)   \\
    - Theorems &    1,714 (30.8\%)  \\
    - Formula &      607 (10.9\%)  \\
    \midrule
    Unique geometric principle & 148 \\
    - Unique Definitions & 65 (43.9\%)  \\
    - Unique Theorems &    47 (31.8\%)   \\
    - Unique Formula &     36 (24.3\%)   \\
    \midrule
    Average question length & 23.4 \\
    Average answer length  & 2.3 \\
    \bottomrule
    \end{tabular}%
    }
    \caption{Key Statistics of our GeoSense.}
    \vspace{-0.5cm}
  \label{tab:sta}%
\end{table}%
\begin{figure}[htbp]
  \centering
  \includegraphics[width=\linewidth]{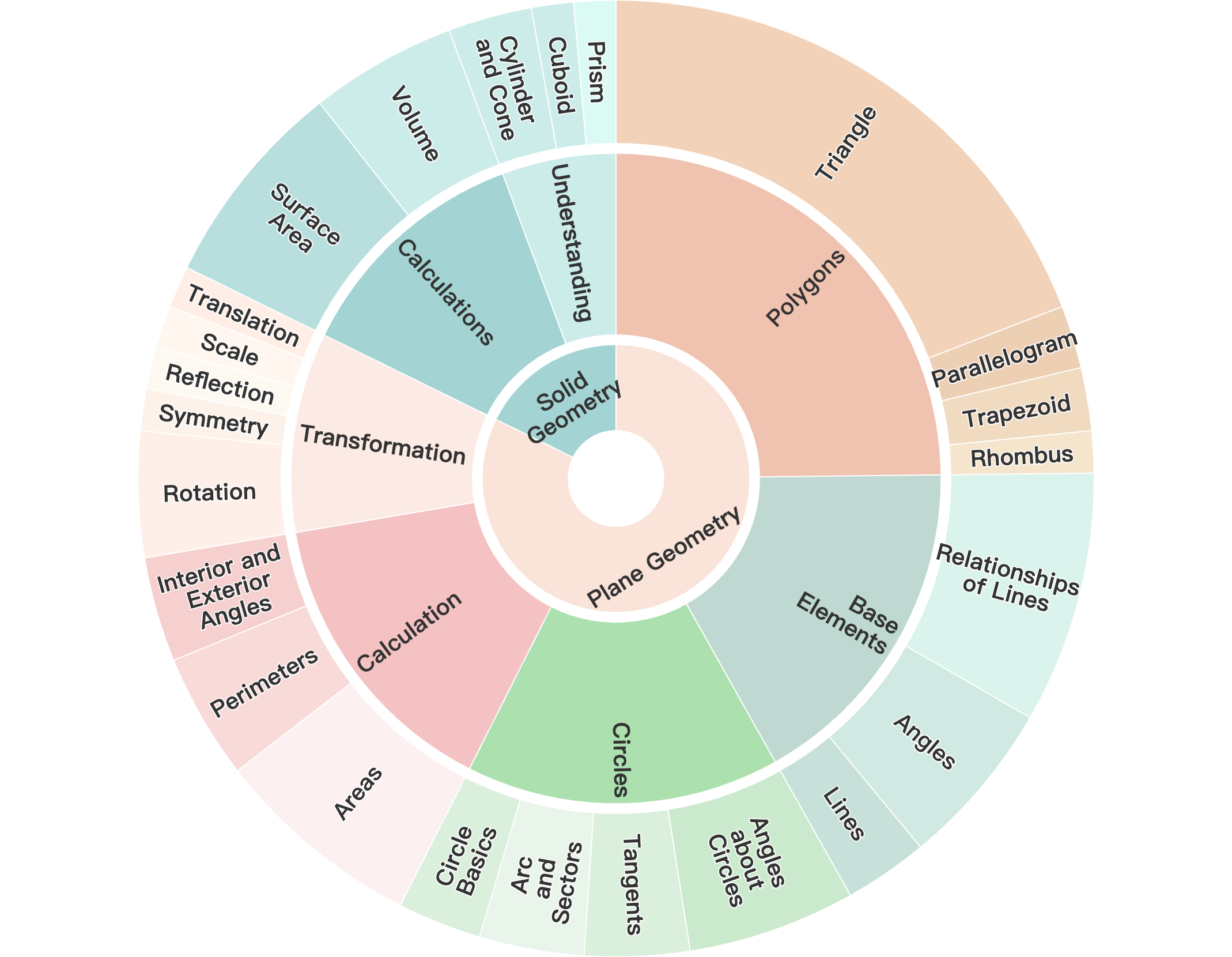}
  \caption{Diagram of the top-3 levels of geometric principles (5 levels in total). See details in Appendix 4.1.}
  \label{fig:knowledge}
\end{figure}
We present the statistical information of our proposed GeoSense dataset in Table~\ref{tab:sta}.
GeoSense consists of 1,789 geometric math problems: 1,116 problems from existing benchmarks~\cite{seo-etal-2015-solving, chen2022geoqageometricquestionanswering, lu2021intergpsinterpretablegeometryproblem, zhang2024mathversedoesmultimodalllm} and 666 problems from the IXL website \footnote{https://www.ixl.com/math/geometry}.
With these geometric math problems, we annotate 5,556 necessary geometric principles (3,235 definitions, 1,714 theorems, and 607 formulas), and their contextual applications.
Furthermore, all geometric principles are categorized into five hierarchical levels, forming a comprehensive framework for geometric knowledge (see Fig.~\ref{fig:knowledge} and details in Appendix 4.1).
We initially categorize these principles into plane and solid geometry, and then progressively refine the classification from various perspectives, such as graphical computation and understanding. This process ultimately leads to detailed definitions, theorems, and formulas. 
With this framework, we could evaluate the MLLMs' mastery of geometric knowledge across different dimensions and granularities for GPS tasks.
\subsection{Data Annotation and Review}
\label{sec:2.2}
We develop a rigorous and meticulous semi-automated annotation pipeline to label the necessary geometric principles and their contextual applications within visual geometric images for each problem. 
Initially, we prompt GPT-4o to generate a detailed geometric problem-solving process for each question, which explicitly specifies the names of the geometric principles and their application. 
Human annotation experts then review and correct any errors in the responses provided by GPT-4o. 
After obtaining a standardized reasoning process, we prompt GPT-4o again to extract the geometric principles and their specific application\footnote{All prompts used during the annotation process are listed in the Supplementary Material.}.
Human annotation experts then standardize the names of the extracted geometric principles and their applications within the problems, ensuring accuracy and consistency. 
Finally, human annotation experts use \texttt{<note></note>} tags to mark the key points in the application process of geometric principles, providing references for subsequent evaluations. 
Notably, our annotation team consists of 23 expert annotators, each holding at least a bachelor's degree. To ensure data quality, the annotation results are cross-validated a minimum of two times. The complete annotation training and the entire annotation process together spanned a period of 4 weeks.


\section{Evaluation Strategy}
\begin{figure*}[t]
  \centering
  \includegraphics[width=\linewidth]{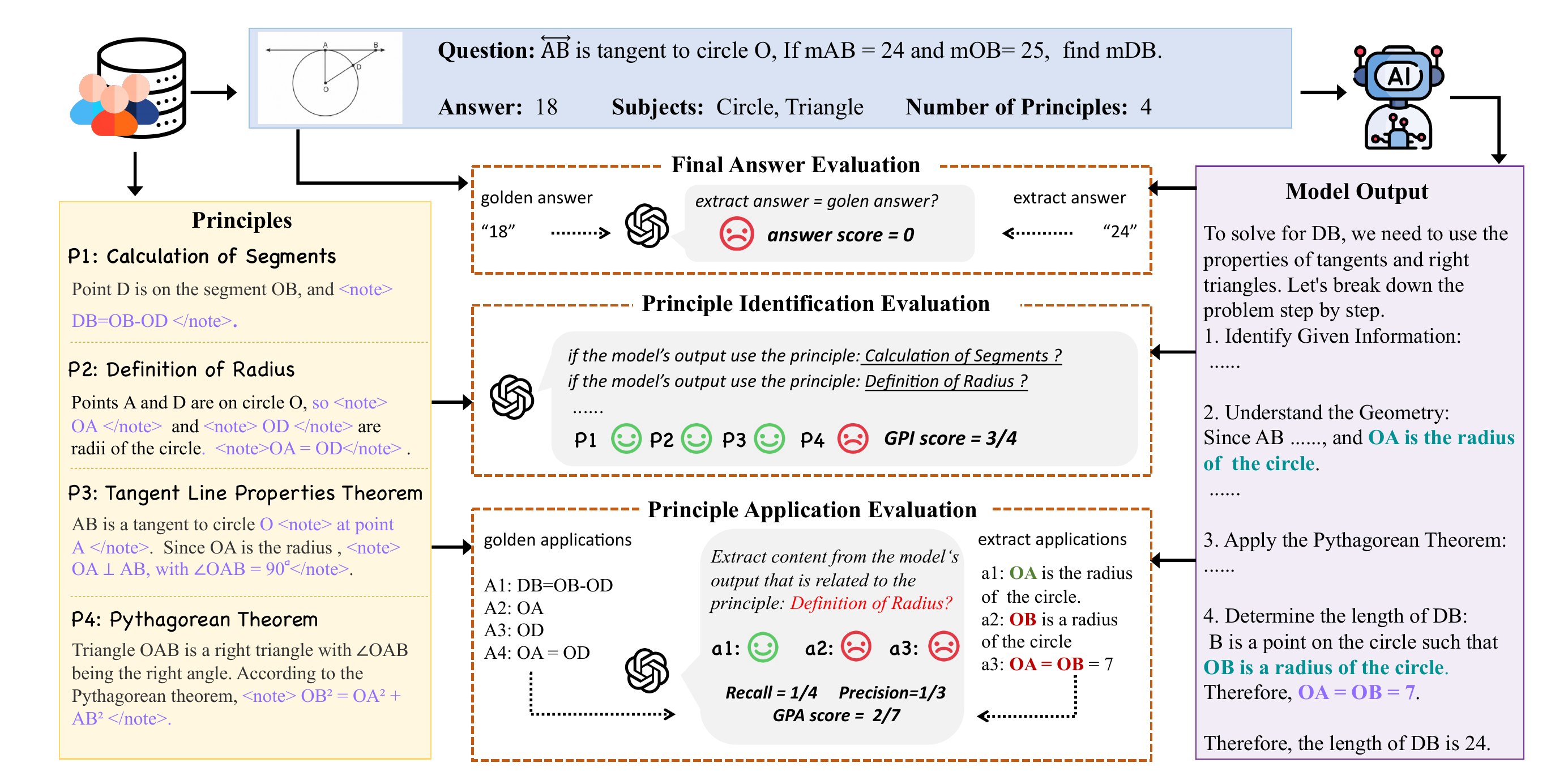}
  \caption{Illustration of GenSense evaluation strategy. MLLMs are assessed through three aspects: identification (i.e., GPI), applications (i.e., GPA) of geometric principles, and final answer accuracy.}
  \label{fig:eval}
\end{figure*}
Geometry problem-solving (GPS) serves as an effective measurement of MLLMs' reasoning capability, requiring both appropriate identification of geometric principles and correct application within complex visual context.
However, existing benchmarks fail to evaluate both aspects, limiting a systematic evaluation of MLLMs on GPS.
In this section, we propose a new evaluation system, consisting of Geometric Principles Identification (GPI) score (Section~\ref{sec:3.1}) and Geometric Principles Application (GPA) score (Section~\ref{sec:3.2}), to fully evaluate human-like reasoning mechanism of MLLMs in GPS. Fig.~\ref{fig:eval} illustrates the overall evaluation framework of GeoSense.



\subsection{Geometric Principles Identification} \label{sec:3.1}
In this section, we propose a Geometric Principles Identification (GPI) score to assess a MLLM's geometric principle identification ability, \emph{i.e.}, whether the MLLM can accurately identify necessary geometric principles to solve the problem.
As shown in Figure~\ref{fig:eval}, given a geometric problem \( q_i \), its corresponding annotation in our dataset explicitly presents the set of necessary geometric principles \( P_i = \{ p_i^1, p_i^2, \ldots, p_i^n \} \). 
For each principle \( p_i^j \), we prompt GPT-4o to determine whether it appears in the MLLM's response. 
The GPI score \( S_i^I \) is defined by:
\begin{equation}
    S_i^I = \frac{\sum_{j=1}^{n} F(p_i^j)}{n},
\end{equation}
where \( F(p_i^j) \in \{0, 1\} \) indicates whether the principle \( p_i^j \) is applied in the MLLM's response. 
The score \( S_i^R \) reflects whether the MLLM can correctly retrieve the appropriate geometric principles when resolving geometric problems. 
However, the GPI score alone can not reveal whether the model applies these principles appropriately within the visual context. 
Such a challenge mirrors the limitations in human geometric reasoning, wherein we may grasp accurate principles such as similar triangles yet still struggle to correctly apply it within the geometric diagram.

\subsection{Geometric Principles Application} \label{sec:3.2}
To evaluate and understand more granularly whether MLLMs can proficiently apply geometric principles to solve problems in geometric reasoning, we further analyze the model's accuracy in applying these principles within a visual context. In our benchmark, we annotate not only the set of geometric principles required for each problem \( P_i = \{ p_i^1, p_i^2, \ldots, p_i^n \} \), but also their corresponding representations in the geometric diagrams \( C_i = \{ c_i^1, c_i^2, \ldots, c_i^n \} \).

We use \texttt{<note>} tags to mark the key elements essential for correct problem-solving, as shown in Figure 1. The key elements within each geometric principle \( c_i^j \) are denoted as \( m_i^j \). When \( F(p_i^j) = 1 \), we use GPT-4o to extract content related to this principle from the model's response \(\hat{y}_i\), denoted as \(\hat{c}_i^j\), with its key elements represented as \(\hat{m}_i^j\). Next, we compare the key elements in \( m_i^j \) and \(\hat{m}_i^j\) to calculate precision and recall for each geometric principle and subsequently compute the F1 score. Based on this, we aggregate the results to obtain the overall knowledge application evaluation measure \( S^A_i \):
\begin{equation}
    S^A_i = \frac{\sum_{j=1}^{n} F(p_i^j) \cdot \frac{2 \times |\hat{m}_i^j \cap m_i^j|}{|\hat{m}_i^j| + |m_i^j|}}{\sum_{j=1}^{n} F(p_i^j)}.
\end{equation}
This measure allows us to effectively evaluate the model's ability to apply geometric principles appropriately within a visual context.

\subsection{Final Answer Evaluation}
We also evaluate the correctness of the final answers. Specifically, given a problem \( q_i \) and its corresponding geometric image \( x_i \), the final answer generated by the model is denoted as \(\hat{y}_i\). We compare \(\hat{y}_i\) with the ground truth \( y_i \) and define the final answer score \( S^{F}_i \) as follows:
\begin{equation}
S^{F}_i = 
\begin{cases} 
1, & \hat{y}_i = y_i \\
0, &  otherwise 
\end{cases}.
\end{equation}
This metric quantifies the overall performance of MLLMs on GPS. 

\begin{table*}[htbp]
  \centering
  \resizebox{\textwidth}{!}{
    \begin{tabular}{l|ccc|ccc|ccc|cccc}
    \toprule
    \multicolumn{1}{c|}{\multirow{2}[4]{*}{\textbf{Model}}} & \multicolumn{3}{c|}{\textbf{Definitions}} & \multicolumn{3}{c|}{\textbf{Theorems}} & \multicolumn{3}{c|}{\textbf{Formulas}} & \multicolumn{4}{c}{\textbf{ALL}} \\
\cmidrule{2-14}          & GPI  & GPA  & ACC.  & GPI  & GPA  & ACC  & GPI  & GPA  & ACC  & GPI  & GPA  & ACC   & AVG. \\
    \midrule
    \multicolumn{14}{c}{\textbf{Closed-Sourced MLLMs}} \\
    \midrule
    Claude35\_Sonnet & 56.5  & 41.2  & 41.9  & 54.9  & 46.8  & 33.8  & 82.8  & 52.5  & 52.9  & 63.2  & 40.8  & 46.1  & 50.0  \\
    Claude37\_Sonnet & 62.0  & 46.7  & 54.3  & 60.2  & 50.0  & 46.5  & \highlight{LightPink}{92.4}  & 56.1  & 67.9  & 68.7  & 45.2  & 57.6  & 57.2  \\
    Gemini-1.5-pro-flash & 60.2  & 43.8  & 53.0  & 58.7  & 51.5  & 45.6  & 85.9 & 55.3  & 56.1  & 67.9  & 44.9  & 55.7  & 56.2  \\
    Gemini-2.0-pro-flash & \highlight{LightPink}{64.2}  & \highlight{LightPink}{47.0}  & \highlight{LightPink}{73.3}  & \highlight{LightPink}{72.7}  & \highlight{LightPink}{59.0}  & \highlight{LightPink}{72.4}  & 87.4  & \highlight{LightPink}{60.0}  & \highlight{LightPink}{77.9}  & \highlight{LightPink}{72.1}  & \highlight{LightPink}{49.7}  & \highlight{LightPink}{74.1}  & \highlight{LightPink}{65.3}  \\
    GPT-4o & 56.3  & 46.3  & 48.0  & 54.1  & 49.3  & 37.4  & 90.8  & 58.3  & 61.1  & 64.4  & 45.3  & 51.7  & 53.8  \\
    \midrule
    \multicolumn{14}{c}{\textbf{Open-Soured MLLMs}} \\
    \midrule
    InternVL2.5-8B & 43.7  & 40.7  & 21.2  & 38.2  & 39.3  & 20.0  & 67.6  & 24.7  & 13.7  & 46.7  & 27.3  & 21.1  & 31.7  \\
    InternVL2.5-38B & 48.7  & 40.6  & 28.9  & 44.5  & 43.9  & 29.8  & 74.8  & 26.4  & 16.0  & 52.7  & 31.1  & 27.3  & 37.0  \\
    InternVL2.5-38B-MPO$^\dagger$ & 50.7  & 44.6  & 29.7  & 48.2  & 46.4  & 30.0  & 75.6  & 29.3  & 16.0  & 53.9  & 33.6  & 27.7  & 38.4  \\
    InternVL2.5-78B & 49.0  & 45.2  & 29.8  & 48.6  & 46.8  & 32.0  & 80.2  & 30.5  & 18.3  & 53.7  & 32.9  & 28.7  & 38.4  \\
\cmidrule{1-14}    
    Deepseek-VL2-small & 25.6  & 35.7  & 23.3  & 26.7  & 36.1  & 19.5  & 67.9  & 48.1  & 30.2  & 34.2  & 23.8  & 26.3  & 28.1  \\
    Deepseek-VL2 & 40.1  & 37.8  & 33.1  & 40.6  & 39.6  & 26.0  & 76.3  & 52.8  & 42.4  & 48.4  & 33.4  & 35.7  & 39.2  \\
\cmidrule{1-14}    
Llama-vision-11B & 43.2  & 36.1  & 22.6  & 37.9  & 35.6  & 18.7  & 74.8  & 37.5  & 29.8  & 47.9  & 29.2  & 24.8  & 34.0  \\
    Llama-vision-90B & 49.1  & 39.2  & 27.3  & 42.0  & 36.0  & 21.2  & 78.2  & 43.6  & 37.0  & 52.9  & 31.4  & 29.8  & 38.0  \\
\cmidrule{1-14} 
    LLaVA-onevison-7B & 36.3  & 38.0  & 22.7  & 39.2  & 39.2  & 22.7  & 72.9  & 40.6  & 42.6  & 41.4  & 26.0  & 22.8  & 30.1  \\
    LLaVA-onevison-72B & 47.9  & 39.0  & 33.7  & 49.6  & 44.8  & 36.4  & 68.3  & 55.9  & 43.1  & 52.5  & 33.2  & 37.2  & 41.0  \\
    \cmidrule{1-14}
    Qwen2-VL-72B & 57.2  & 44.2  & 46.6  & 57.7  & 44.2  & 46.6  & 85.5  & 52.0  & 50.4  & 64.0  & 43.4  & 49.2  & 52.2  \\
    Qwen2.5-VL-3B & 50.5  & 39.9  & 33.5  & 48.8  & 47.0  & 27.7  & 74.8  & 45.0  & 41.2  & 55.2  & 36.5  & 34.9  & 42.2  \\
    Qwen2.5-VL-7B & 57.7  & 45.6  & 43.6  & 57.4  & 51.2  & 37.5  & 85.9  & 60.4  & 53.1  & 63.1  & 44.6  & 46.3  & 51.3  \\
    Qwen2.5-VL-72B & 61.5  & 47.5  & \highlight{LightCyan}{61.5}  & \highlight{LightCyan}{65.1}  & 54.8  & \highlight{LightCyan}{57.5}  & \highlight{LightCyan}{89.7}  & \highlight{LightCyan}{61.5}  & \highlight{LightCyan}{63.8}  & 68.5  & 48.1  & \highlight{LightCyan}{63.8}  & \highlight{LightCyan}{60.1}  \\
    QVQ-72B-Preview$^\dagger$ & \highlight{LightCyan}{68.2}  & \highlight{LightCyan}{56.0}  & 53.1  & 63.6  & \highlight{LightCyan}{58.3}  & 49.6  & 85.1  & 58.4  & 54.2  & \highlight{LightCyan}{72.3}  & \highlight{LightCyan}{53.5}  & 54.3  & 60.0  \\
    \bottomrule
    \end{tabular}%
    }
  \caption{Mathematical Evaluation on Different Types of Geometric Principles in GeoSense. GPI refers to Geometric Principles Identification, GPA means Geometric Principles Application. MLLMs with $\dagger$ are particularly adept at reasoning tasks.}
  \vspace{-0.7cm}
  \label{tab:main_results}%
\end{table*}

\begin{table*}[htbp]
  \centering
  \resizebox{\textwidth}{!}{
    \begin{tabular}{l|ccc|ccc|ccc|ccc|ccc}
    \toprule
    \multicolumn{1}{c|}{\multirow{2}[4]{*}{\textbf{Model}}} & \multicolumn{3}{c|}{\textbf{CSF}} & \multicolumn{3}{c|}{\textbf{USF}} & \multicolumn{3}{c|}{\textbf{TMPF}} & \multicolumn{3}{c|}{\textbf{CPF}} & \multicolumn{3}{c}{\textbf{UPF}} \\
\cmidrule{2-16}          & GPI  & GPA  & ACC  & GPI  & GPA  & ACC  & GPI  & GPA  & ACC  & GPI  & GPA  & ACC   & GPI  & GPA  & ACC \\
    \midrule
    \multicolumn{16}{c}{\textbf{Closed-Sourced MLLMs}} \\
    \midrule
    Claude35\_Sonnet & 85.0  & 53.5  & 53.8  & 80.9  & 34.7  & 54.6  & 65.9  & 32.5  & 27.2  & 68.4  & 60.5  & 56.1  & 45.1  & 38.7  & 40.7  \\
    Claude37\_Sonnet & 91.1  & 62.5  & 76.8  & 80.8  & 36.5  & 73.5  & 82.7  & 39.4  & 47.3  & 70.5  & 63.5  & 68.5  & 54.1  & 44.8  & 52.0  \\
    Gemini-1.5-pro-flash & 87.3  & 62.6  & 62.1 & \highlight{LightPink}{86.0}  & 33.5  & 64.2  & {83.9}  & 44.1  & 48.4  & 73.1  & 65.4  & 64.8  & 50.7  & 43.5  & 49.3  \\
    Gemini-2.0-pro-flash & 88.1  & 58.8  & \highlight{LightPink}{89.9}  & 72.4  & 35.1  & \highlight{LightPink}{89.9}  & \highlight{LightPink}{84.6}  & \highlight{LightPink}{47.2}  & \highlight{LightPink}{62.1}  & \highlight{LightPink}{77.7}  & 69.5  & \highlight{LightPink}{77.0}  & \highlight{LightPink}{60.3}  & \highlight{LightPink}{51.2}  & \highlight{LightPink}{70.9}  \\
    GPT-4o & \highlight{LightPink}{91.3}  & \highlight{LightPink}{66.8}  & 72.3  & 73.7  & \highlight{LightPink}{37.0}  & 75.1  & 83.5  & 37.0  & 34.7  & 72.7  & \highlight{LightPink}{73.1}  & 66.1  & 49.4  & 44.8  & 44.2  \\
    \midrule
    \multicolumn{16}{c}{\textbf{Open-Soured MLLMs}} \\
    \midrule
    InterVL2.5-8B & 74.3  & 28.7  & 14.2  & 55.0  & 28.5  & 21.2  & 73.6  & 31.9  & 17.0  & 55.9  & 36.3  & 29.1  & 35.0  & 35.3  & 21.5  \\
    InterVL2.5-38B & 82.4  & 35.3  & 17.1  & 67.0  & 32.9  & 24.1  & 82.7  & 28.0  & 23.2  & 57.5  & 43.3  & 37.4  & 39.8  & 36.7  & 28.6  \\
    InterVL2.5-38B-MPO$^\dagger$ & 84.0  & 34.8  & 16.4  & 62.1  & 33.5  & 23.3  & 87.9  & 35.7  & 26.1  & 56.3  & 42.8  & 35.6  & 41.9  & 39.7  & 30.1  \\
    InterVL2.5-78B & \highlight{LightCyan}{90.1}  & 34.5  & 17.4  & 65.0  & 35.4  & 22.5  & 86.0  & 34.8  & 27.6  & 61.5  & 46.0  & 36.2  & 40.2  & 41.7  & 30.8  \\
    \midrule
    Deepseek-VL2-small & 66.3  & 51.8  & 34.1  & 52.0  & 25.3  & 38.9  & 53.7  & 22.3  & 16.3  & 47.0  & 59.7  & 40.1  & 21.9  & 28.8  & 20.0  \\
    Deepseek-VL2 & 79.4  & 55.0  & 49.2  & 53.7  & 40.3  & 51.7  & 49.0  & 33.6  & 30.4  & 58.6  & 56.0  & 48.3  & 33.6  & 35.6  & 28.6  \\
    \midrule
    Llama-vision-11B & 77.9  & 41.0  & 33.9  & 55.0  & 33.4  & 37.3  & 58.9  & 19.1  & 14.8  & 52.6  & 46.4  & 42.7  & 32.2  & 33.7  & 20.7  \\
    Llama-vision-90B & 83.4  & 52.3  & 45.8  & 70.5  & 32.7  & 45.1  & 68.2  & 21.5  & 19.7  & 58.8  & 52.3  & 44.3  & 37.5  & 34.8  & 24.1  \\
    \midrule
    LLaVA-onevison-7B & 79.3  & 42.6  & 32.8  & 57.3  & 26.2  & 35.3  & 52.3  & 22.4  & 16.0  & 51.6  & 43.8  & 32.2  & 28.6  & 33.6  & 21.7  \\
    LLaVA-onevison-72B & 65.7  & 65.5  & 50.0  & 48.1  & 35.3  & 56.3  & 71.4  & 19.3  & 24.8 & 65.5  & 54.2  & 43.3  & 39.1  & 37.7  & 35.0  \\
    \midrule
    Qwen2-VL-72B & 82.5  & 57.8  & 57.4  & 70.7  & 41.2  & 67.1  & 78.6  & 29.8  & 22.2  & 72.1  & 61.0  & 67.1  & 49.1  & 43.2  & 43.7  \\
    Qwen2.5-VL-3B & 77.6  & 48.9  & 48.7  & 68.4  & 34.3  & 60.0  & 73.4  & 26.8  & 17.4  & 66.1  & 57.1  & 53.1  & 40.6  & 39.5  & 29.5  \\
    Qwen2.5-VL-7B & 85.9  & 63.9  & 59.6  & 72.9  & 43.1  & 67.9  & 82.3  & 36.0  & 26.3  & 67.2  & 71.1  & 60.3  & 48.9  & 44.5  & 40.6  \\
    Qwen2.5-VL-72B & 88.2  & 64.4  & \highlight{LightCyan}{74.1}  & 73.9  & 36.6  & \highlight{LightCyan}{78.7}  & \highlight{LightCyan}{89.7}  & 37.6  & 42.0  & \highlight{LightCyan}{76.1}  & 69.7  & \highlight{LightCyan}{77.6}  & 55.2  & 47.3  & \highlight{LightCyan}{56.7}  \\
    QVQ-72B-Preview$^\dagger$ & 87.3  & \highlight{LightCyan}{71.1}  & 66.2  & \highlight{LightCyan}{76.7}  & \highlight{LightCyan}{45.8}  & 67.9  & 82.5  & \highlight{LightCyan}{42.1}  & \highlight{LightCyan}{47.4}  & 74.0  & \highlight{LightCyan}{71.2}  & 65.3  & \highlight{LightCyan}{58.5}  & \highlight{LightCyan}{52.9}  & 48.2  \\
    \bottomrule
    \end{tabular}%
    }
  \caption{Mathematical Evaluation on Different Subjects in GeoSense. GPI = Geometric Principles Identification, GPA= Geometric Principles Application, Calculation of Solid Figures = CSF, Understanding of Solid Figures = USF, Transformation and Motion of Plane Figures = TMPF, Calculation of Plane Figures = CPF and Understanding of Plane Figures = UPF. MLLMs with $\dagger$ are typically trained for reasoning tasks.}
  \vspace{-0.7cm}
  \label{tab:main_results2}%
\end{table*}

\section{Experiments}
In this section, we systematically evaluate existing MLLMs on GeoSense. First, we introduce the experimental setup in Section~\ref{sec:expset}. Then, we detail the experimental results and analysis in Section~\ref{sec:expana}, and present the error analysis in Section~\ref{sec:experr}.
\subsection{Experimental Setup}
\label{sec:expset}
\paragraph{\textbf{Evaluation Models.}} We evaluate the geometric reasoning capabilities of three categories of MLLMs on GeoSense: 1) closed-source MLLMs, including Claude35\_Sonnet\footnote{https://www.anthropic.com/news/claude-3-5-sonnet}, Claude37\_Sonnet\footnote{https://www.anthropic.com/news/claude-3-7-sonnet}, Gemini-2.0-pro-flash\footnote{https://deepmind.google/technologies/gemini/flash/}, and GPT-4o\footnote{https://openai.com/index/}; 2) open-source MLLMs, including the InternVL2.5 series \cite{chen2024internvlscalingvisionfoundation}, Deepseek-VL2 \cite{wu2024deepseekvl2mixtureofexpertsvisionlanguagemodels} series, Qwen2-VL series \cite{wang2024qwen2vlenhancingvisionlanguagemodels}, Qwen2.5-VL\footnote{https://help.aliyun.com/zh/model-studio/developer-reference/use-qwen-by-calling-api} series, Llama-vision-11B/90B\footnote{https://ollama.com/library/llama3.2-vision}, and LLaVA-onevision-0.6B/72B \cite{li2024llavaonevisioneasyvisualtask}; and 3) reasoning MLLMs, including InternVL2.5-38B-MPO \cite{wang2024enhancingreasoningabilitymultimodal}, QVQ-72B-Preview\footnote{https://qwenlm.github.io/zh/blog/qvq-72b-preview/}.

\paragraph{\textbf{Implementation Details.}}
All our evaluations adopt CoT prompting technique\cite{wei2023chainofthoughtpromptingelicitsreasoning}. Additionally, the models are further required to explicitly identify and apply the necessary geometric principles during the reasoning steps. All experiments are conducted in a zero-shot setting to reveal the models' general reasoning abilities. Open-source model reasoning is performed on NVIDIA A100 GPUs, while closed-source model reasoning is conducted via their official API calls. The temperature and sampling parameters are set to the official default settings for each model. For evaluation, we use ``GPT-4o-0513'' as the judge.
\subsection{Experimental results}
\label{sec:expana}

To systematically examine the multimodal reasoning capabilities grounded in geometric principles, we report the evaluation results of various models on GeoSense for three attributes (i.e., definitions, theorems, and formulas) as depicted in Table~\ref{tab:main_results}. Additionally, we measure five detailed topics namely, calculation of solid figures (CSF), understanding of solid figures (USF), transformation and motion of plane figures (TMPF), calculation of plane figures (CPF), and understanding of plane figures (UPF) as illustrated in Table~\ref{tab:main_results2}. 

From the results, we notice that closed-sourced models generally outperform open-source models. Among them, Gemini-2.0-Pro-Flash performs the best, with an average score of $65.3$. MLLMs trained specifically for reasoning tasks show significant improvements in reasoning abilities; for example, InternVL2.5-38B-MPO exhibits a $9.1\%$ average performance improvement (AVG) compared to InternVL2.5-38B. QVQ-72B-Preview performs well in GPI and GPA but has a lower ACC than Gemini-2.0-Pro-Flash, primarily due to overthinking leading to incorrect answers. Additionally, larger model sizes contribute to performance enhancements, as seen in the Qwen2.5-VL series, where overall performance improves with increasing model size. 

From the results across different subjects in Table~\ref{tab:main_results2} and Fig.~\ref{fig}, we observe that models perform better in solid geometry than in plane geometry, and understanding plane geometric figures is a common weakness among all MLLMs. This is due to the large number of geometric principles involved in plane geometry, which includes many easily confused concepts (such as determining the similarity and congruence of triangles). These factors pose greater challenges to MLLMs in processing visual information and reasoning about spatial relationships.
\begin{figure}[t]
  \centering
  \begin{subfigure}[b]{0.48\linewidth}
    \centering
    \includegraphics[width=\linewidth]{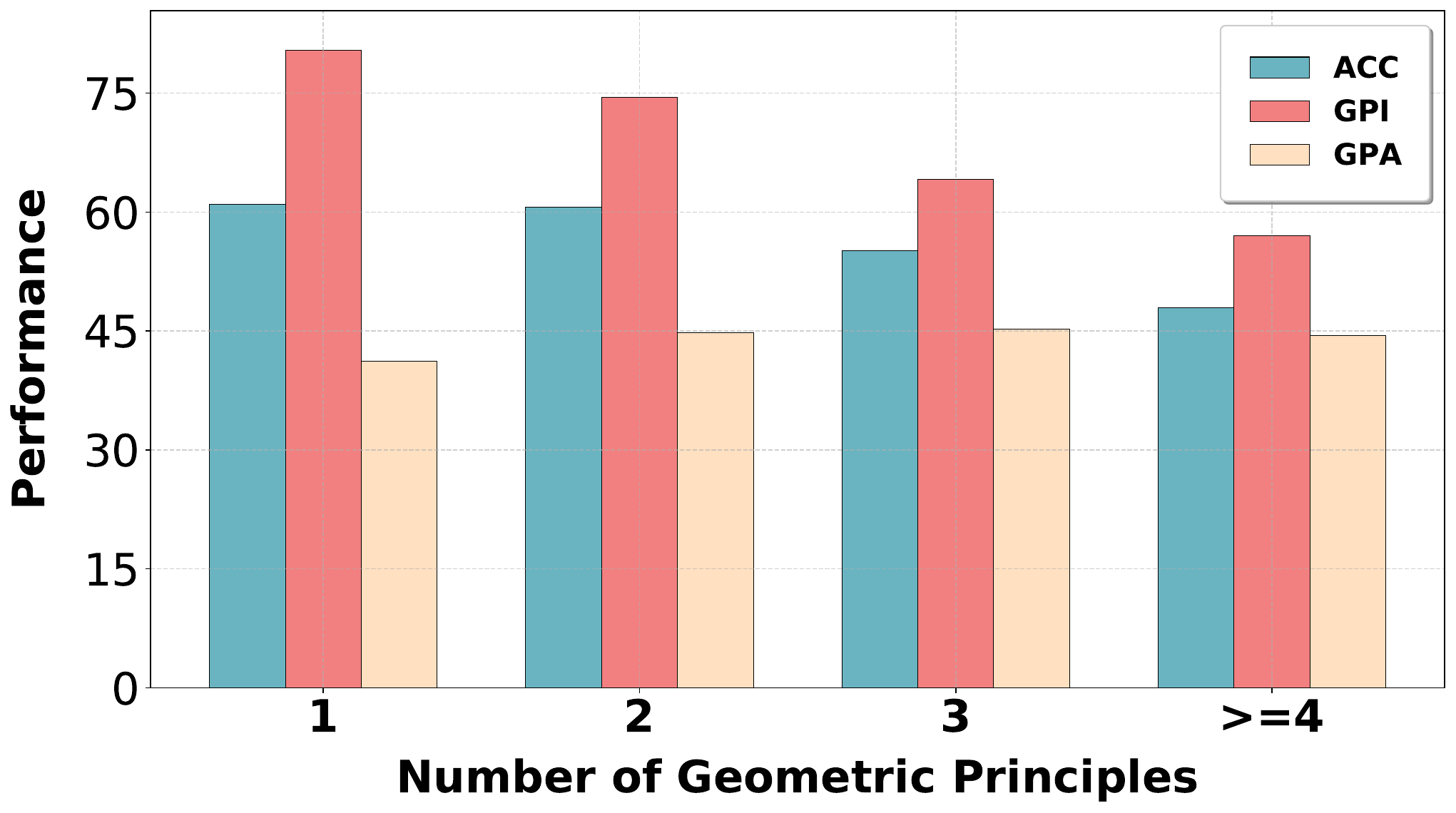}
    \caption{Closed-sourced MLLMs.}
    \Description{A woman and a girl in white dresses sit in an open car.}
    \label{fig:num1}
  \end{subfigure}
  \begin{subfigure}[b]{0.48\linewidth}
    \centering
    \includegraphics[width=\linewidth]{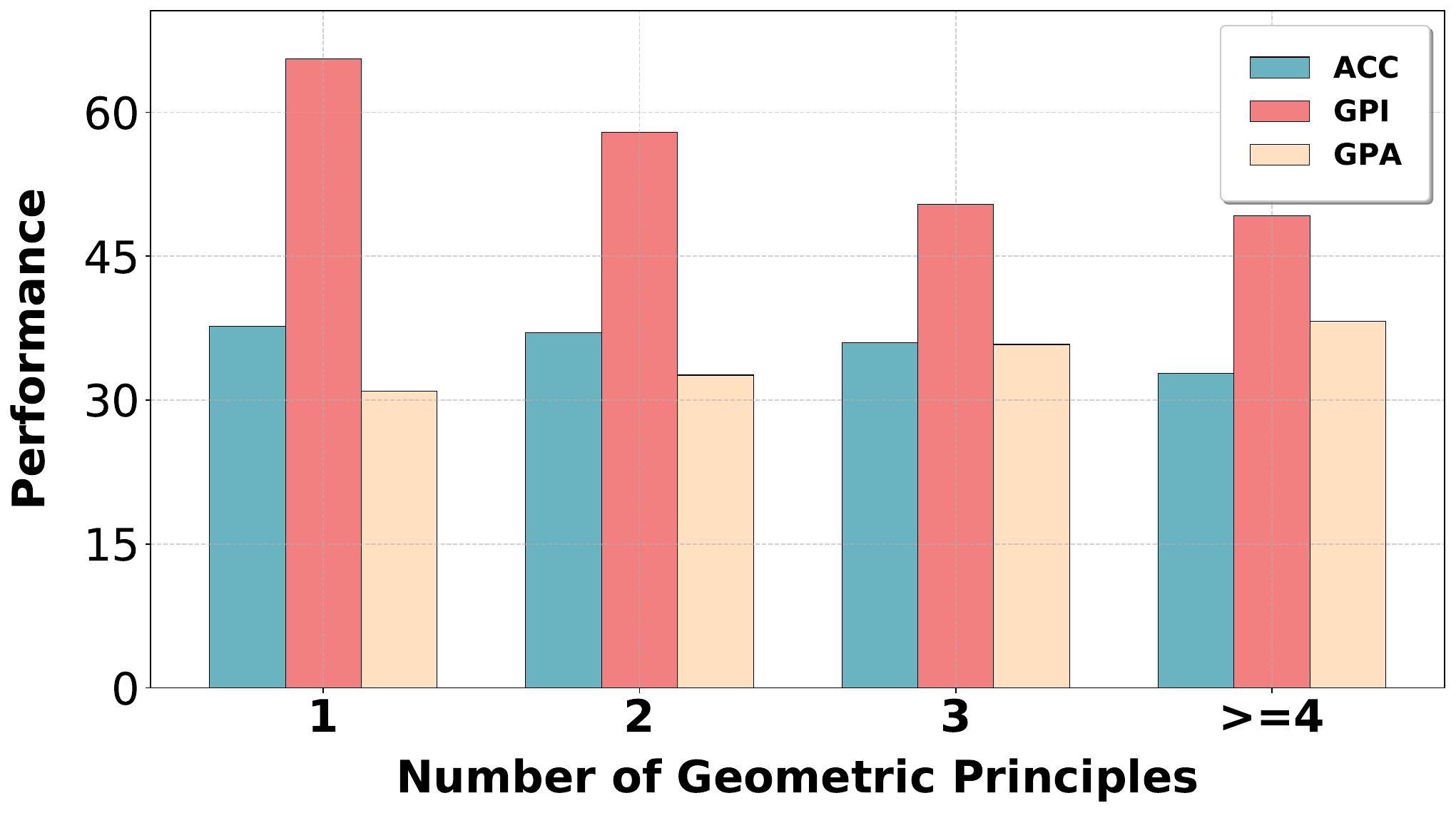}
    \caption{Open-sourced MLLMs.}
    \Description{A woman and a girl in white dresses sit in an open car.}
    \label{fig:num2}
  \end{subfigure}
  \vspace{-0.5cm}
  \caption{The performance of (a) Closed-sourced and (b) Open-sourced MLLMs on problems with different number of geometric principles.}
  \vspace{-0.5cm}
  \label{fig:num_ana}
\end{figure}


\subsection{Experimental Analysis}
We primarily analyze performance of MLLMs on GeoSense, and have the following observations:

\textbf{GPI and GPA jointly affect MLLMs' reasoning abilities.}
MLLMs' ACC scores on GeoSense are concurrently determined by both GPI and GPA scores.
In Table~\ref{tab:main_results}, InternVL-2.5-8B and InternVL-2.5-38B achieve similar GPA scores for retrieving geometric principles.
However, InternVL-2.5-38B achieves ACC $7.7\%$ higher than InternVL-2.5-8B because of its $5\%$ higher GPI score.
Additionally, more examples, such as Claude37\_Sonnet and Qwen2.5-VL-72B, further demonstrate that a decrease in GPA also impedes model performance under full-scale evaluation. 
Across different subjects in Table~\ref{tab:main_results2}, most MLLMs show minimal variation in GPI scores between CPF and TMPF tasks.
However, the ACC scores of MLLMs tend to decrease in TMPF due to lower GPA scores.
Moreover, most existing MLLMs show relatively limited GPI and GPA scores on GeoSense, indicating that the identification accuracy and application correctness of geometric principles jointly limit the reasoning ability of MLLMs on GPS.

\textbf{Why MLLMs perform worse on complex problems.}
Intuitively, more complex geometric problems require more geometric principles.
In Figure~\ref{fig:num_ana}, we exhibit how our proposed metrics vary with the complexity of geometric problems.
We utilize the average scores of open-sourced and closed-sourced models to represent MLLMs' performance and the numbers of geometric principles to represent the complexity.
We observe that both GPI and ACC scores decrease as the complexity increases, while GPA scores show a negligible impact.
Such a trend is even more evident in closed-source MLLMs.
These observations suggest that MLLMs' worse performances on complex problems are mainly caused by the failure to accurately identify essential geometric principles.
This experiment highlights the importance of improving the MLLMs' ability to identify geometric principles more accurately to further enhance their reasoning capabilities.

\textbf{MLLMs excel in computation but fail in understanding.} From the results in Table~\ref{tab:main_results}, the performance of the three metrics for MLLMs under Formulas is significantly higher than under Definitions and Theorems, especially for GPI metric. This indicates that MLLMs can more clearly identify the required geometric principles when faced with computational problems. In contrast, definitions and theorems often contain abstract properties and relationships of geometric elements, which MLLMs struggle to understand. 

\begin{figure*}[htbp]
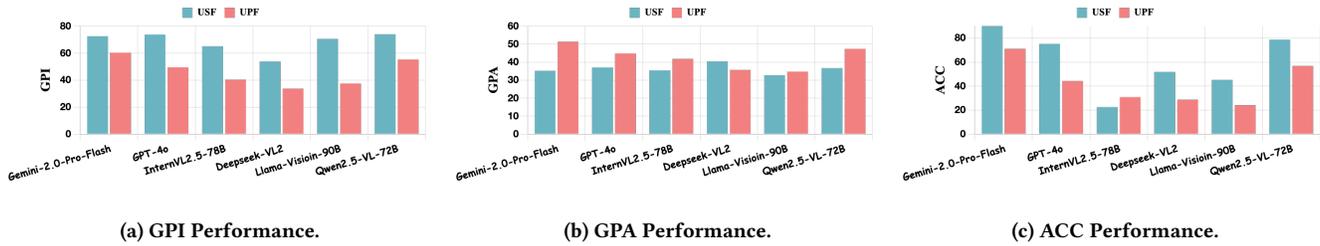

  \centering
  \begin{subfigure}[b]{0.33\linewidth}
    \centering
    \includegraphics[width=\linewidth]{figs/com_gpi.pdf}
    \caption{GPI Performance.}
    \Description{A woman and a girl in white dresses sit in an open car.}
    \label{fig:com1}
  \end{subfigure}
  \begin{subfigure}[b]{0.33\linewidth}
    \centering
    \includegraphics[width=\linewidth]{figs/com_gpa.pdf}
    \caption{GPA Performance.}
    \Description{A woman and a girl in white dresses sit in an open car.}
    \label{fig:com2}
  \end{subfigure}
  \begin{subfigure}[b]{0.33\linewidth}
    \centering
    \includegraphics[width=\linewidth]{figs/com_acc.pdf}
    \caption{ACC Performance.}
    \Description{A woman and a girl in white dresses sit in an open car.}
    \label{fig:com2}
    \end{subfigure}
  \caption{The (a) GPI, (b) GPA, (c) ACC performance of MLLMs on USF and UPF.}
  \label{fig:com_ana}
\end{figure*}

\textbf{GPI limits MLLMs' performance in plane geometry.} Fig.~\ref{fig:com_ana} illustrates the differences in various metrics for several models on USF and UPF. In terms of the GPI metric, we find that most models perform better on the USF subject compared to UPF. However, for the GPA metric, the performance difference between the two subjects is not significant, with models, except for Deepseek-VL, even being stronger in the UPF domain. Nonetheless, regarding the ACC metric, most models still perform better on USF. These observations suggest that the key factor limiting the models' ability to solve plane geometry problems is GPI, i.e., the difficulty models face in accurately identifying the necessary geometric principles. This is due to the numerous and easily confusable principles in plane geometry, such as determining similar and congruent triangles. This highlights the need for models to accurately identify necessary principles to enhance their understanding of plane geometry.

\subsection{Error Analysis}
\label{sec:experr}
\begin{figure}[htbp]
  \centering
  \begin{subfigure}[b]{0.49\linewidth}
    \centering
    \includegraphics[width=\linewidth]{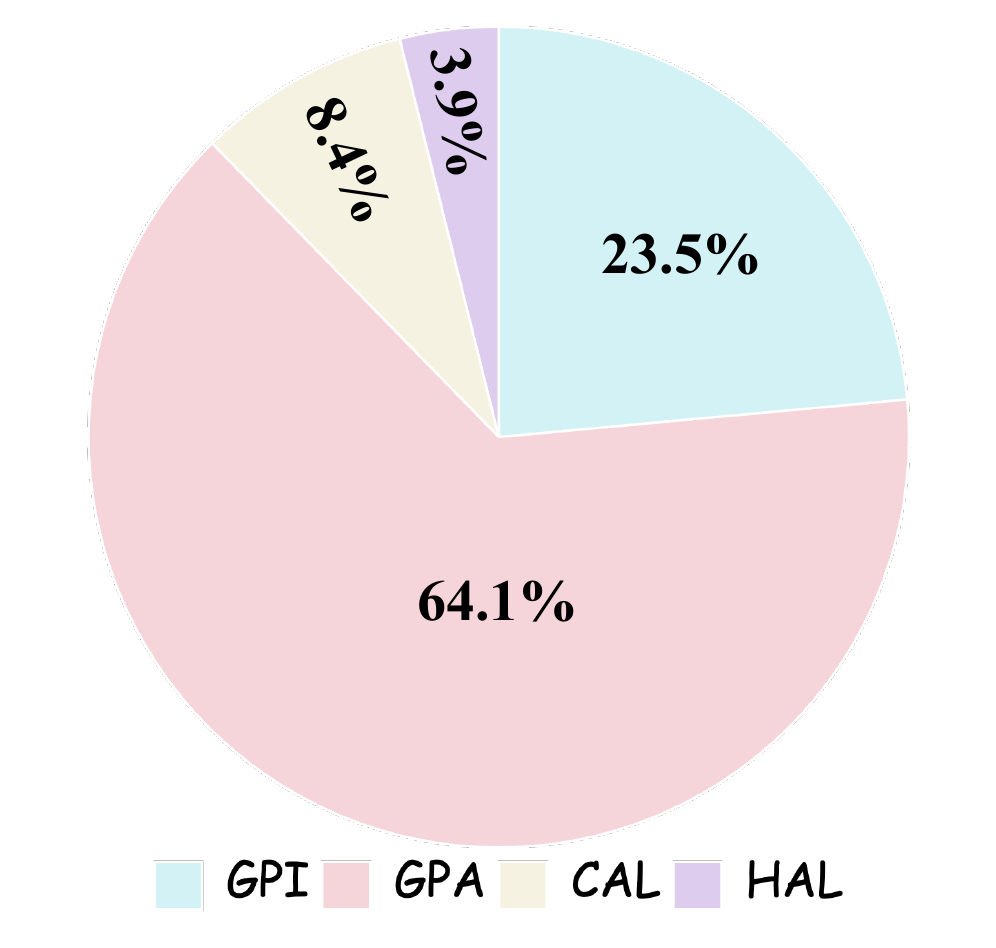}
    \caption{Gemini-2.0-Pro-Flash.}
    \Description{A woman and a girl in white dresses sit in an open car.}
    \label{fig:com1}
  \end{subfigure}
  \begin{subfigure}[b]{0.49\linewidth}
    \centering
    \includegraphics[width=\linewidth]{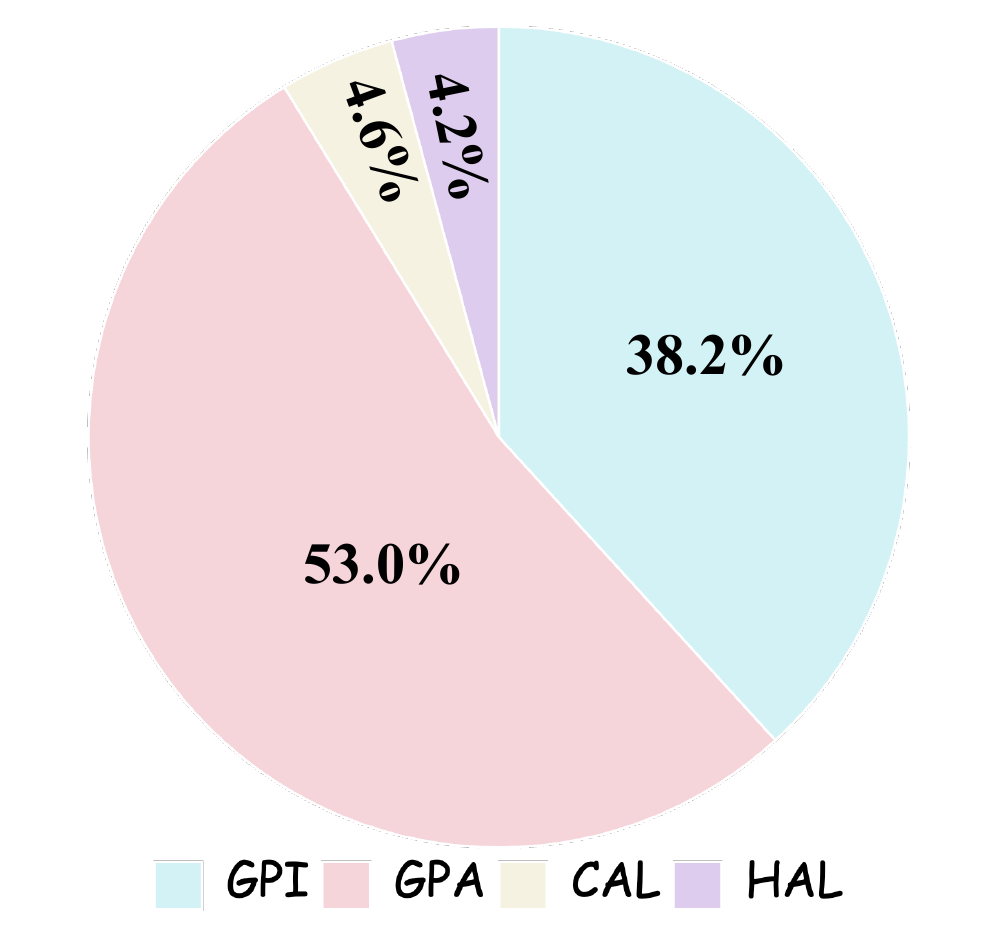}
    \caption{Qwen2.5-VL-72B.}
    \Description{A woman and a girl in white dresses sit in an open car.}
    \label{fig:com2}
  \end{subfigure}
  \caption{Error Analysis of Leading Closed-Source and Open-Source MLLMs.}
  \label{fig:err_ana}
\end{figure}
To gain a deeper insight into the performance bottlenecks of state-of-the-art MLLMs on GPS tasks, we analyze the error distribution of the leading closed-source and open-source models: Gemini-2.0-Pro-Flash and Qwen2.5-VL-72B. For each problem, we identify the critical errors in their reasoning process and categorize these errors into four types: geometric principles identification errors, geometric principles application errors, calculation errors, and hallucinations.

As illustrated in Fig.~\ref{fig:err_ana}, the primary source of errors for the SOTA models is GPA, which involves incorrectly applying geometric principles within visual geometric contexts. Errors in perception and the inability to align geometric principles with visual elements lead to GPA errors. Secondly, GPI represents the second major source of errors, with Gemini-2.0-Pro-Flash and Qwen2.5-VL-72B exhibiting 23.5\% and 38.2\% of errors due to GPI, respectively. This indicates that enhancing the models’ ability to recognize geometric principles could further improve their reasoning capabilities. Additionally, a small number of errors are attributed to calculation errors and model hallucinations, and addressing these issues is crucial for optimizing overall model performance.

%

\section{Related Work}
\subsection{Multi-modal Large Language Models}
The synergistic evolution of Large Language Models (LLMs)\cite{brown2020languagemodelsfewshotlearners,jiang2024mixtralexperts,touvron2023llamaopenefficientfoundation,touvron2023llama2openfoundation,openai2023chatgpt} and vision foundation models\cite{radford2021learningtransferablevisualmodels,kirillov2023segment,zhang2023promptgeneratecachecascade,zhang2023personalizesegmentmodelshot,zhang2022learning3drepresentations2d} has formed the core driving force behind the development of MLLMs. Building upon breakthroughs in text-based models \cite {touvron2023llamaopenefficientfoundation,brown2020languagemodelsfewshotlearners} and vision frameworks \cite{radford2021learningtransferablevisualmodels, kirillov2023segment}, researchers have progressively established fundamental cross-modal interaction capabilities. MLLMs have achieved performance breakthroughs through massive-scale data training, where closed-source models like OpenAI's GPT-4V\cite{openai2023gpt4v} and Google's Gemini\cite{geminiteam2024geminifamilyhighlycapable} have set benchmarks in complex visual reasoning tasks. Concurrently, open-source community initiatives such as LLaVA\cite{liu2023llava,liu2024improvedbaselinesvisualinstruction,liu2024llavanext} and MiniGPT-4\cite{chen2023minigptv2largelanguagemodel,zhu2023minigpt4enhancingvisionlanguageunderstanding} employ frozen CLIP\cite{radford2021learningtransferablevisualmodels} models for image encoding while injecting visual prompts into LLaMA\cite{touvron2023llamaopenefficientfoundation} for multimodal instruction tuning. As technology iterates, multimodal architectures continue to evolve: mPLUG-Owl\cite{ye2024mplugowlmodularizationempowerslarge,ye2023mplugowl2revolutionizingmultimodallarge} proposes cross-modal collaborative training mechanisms, Qwen-VL\cite{bai2023qwenvlversatilevisionlanguagemodel} enhances fine-grained understanding through spatial-aware modules, and InternLM-XComposer\cite{dong2024internlmxcomposer2masteringfreeformtextimage} along with SPHINX\cite{liu2024sphinxxscalingdataparameters,lin2023sphinxjointmixingweights} adopt mixture-of-experts architectures to boost multi-task performance. Recent advancements in large reasoning LLMs like OpenAI o1\cite{openai2024o1} and DeepSeekR1\cite{deepseekai2025deepseekr1incentivizingreasoningcapability}, which demonstrate remarkable progress in complex problem-solving, have spurred numerous explorations to enhance MLLMs' reasoning capabilities\cite{shen2025vlmr1,chen2025r1v}. Specialized algorithms have also been developed to strengthen MLLMs' mathematical and geometric reasoning capacities. In this paper, we introduce the GeoSense benchmark to comprehensively evaluate MLLMs' ability to solve geometric problems by leveraging geometric principles.

\subsection{Geometry Benchmarks}
Mathematical reasoning has become a pivotal area within contemporary AI research, posing significant challenges for LLMs and MLLMs. Initially, datasets in this domain targeted elementary algebra\cite{hendrycks2021measuringmathematicalproblemsolving} and arithmetic word problems\cite{roy2016solvinggeneralarithmeticword}, which were relatively limited in scope and number. Subsequent efforts expanded to more complex and diversified mathematical problem sets, exemplified by datasets like MATH\cite{hendrycks2021measuringmathematicalproblemsolving}, GSM8K\cite{cobbe2021trainingverifierssolvemath}, and MMLU\cite{hendrycks2021measuringmassivemultitasklanguage}. These datasets notably enhanced the difficulty and breadth of mathematical questions, thereby establishing robust benchmarks for evaluating general-purpose and math-specific language models\cite{zhou2023solvingchallengingmathword,yue2023mammothbuildingmathgeneralist,wang2023mathcoderseamlesscodeintegration,gao2023gllavasolvinggeometricproblem,luo2025wizardmathempoweringmathematicalreasoning}. Moreover, the rapid advancement of MLLMs has spurred the need for high-quality multimodal benchmarks to assess the models' capabilities in solving mathematical problems within visually enriched contexts. For instance, datasets such as GeoQA\cite{chen2022geoqageometricquestionanswering}, UniGeo\cite{chen2022unigeounifyinggeometrylogical}, and Geometry3K\cite{lu2021intergpsinterpretablegeometryproblem} focus specifically on geometry-related queries. In addition, initiatives like MathVista\cite{lu2024mathvistaevaluatingmathematicalreasoning} have broadened the scope to incorporate a range of multimodal tasks involving mathematical reasoning, while MMMU\cite{yue2024mmmumassivemultidisciplinemultimodal} addresses college-level problems requiring intricate domain-specific knowledge. These multimodal benchmarks significantly advance the evaluation of models in complex mathematical reasoning and their application across modalities. Nonetheless, current benchmarks in the geometry problem-solving domain still exhibit notable shortcomings, particularly in systematically evaluating the cross-modal application of geometric principles. 
Thus, developing comprehensive and systematic benchmarks that fully assess cross-modal capabilities in geometric reasoning remains an imperative research direction.

\section{Conclusion}
In this paper, we introduce GeoSense, the first comprehensive bilingual benchmark to systematically evaluate the reasoning abilities of MLLMs with a focus on identifying and applying geometric principles. We first establish a comprehensive framework that includes 148 unique geometric principles. Additionally, we curate a dataset that comprises 1,789 geometric math problems, annotate 5,556 geometric principles and their application within geometric images. Moreover, we introduce two novel evaluation metrics GPI and GPA to assess MLLMs’ ability to identify correct geometric principles and apply them to specific elements within geometric diagrams, respectively. Extensive experiments reveal insights into the performance of different MLLMs, highlighting their limitations in applying geometric principles within visual contexts.
\bibliographystyle{ACM-Reference-Format}
\bibliography{main}
\appendix
\section{Overview}
\begin{itemize}
    \item Ethical Concerns  (\S \ref{sm:eth})
    \item Limitation and future work (\S \ref{sm:limitation}) 
    \item More Dataset Details (\S \ref{sm:dataset}) 
        \begin{itemize}
            \item Geometric Principles Structure (\S \ref{sm:cps})
            \item Examples with Geometric Principles
Annotated in GeoSense (\S \ref{sm:egpas})
        \end{itemize}
    \item Additional experimental details (\S \ref{sm:experimental})
    \begin{itemize}
            \item Evaluation Results of More MMLMs on
GeoSense-English (\S \ref{sm:ermge})
            \item Response of Differenet MLLMs (\S \ref{sp:rdm})
            \item Prompts for Evaluation Strategy (\S \ref{sm:pes})
        \end{itemize}
\end{itemize}
\section{Ethical Concerns}
\label{sm:eth}
We guarantee that GeoSense fully complies with legal and ethical standards throughout its data collection and annotation processes, ensuring no violations occur. All annotation experts involved receive fair compensation. The data for GeoSense is derived from publicly accessible test questions and professional websites. Given the standardized nature of answers in mathematical problems, cultural differences do not affect them. Moreover, we confirm that GeoSense is intended exclusively for academic research, with a firm prohibition on any commercial use. In addition, we declare that we will take full responsibility for any infringement of rights and acknowledge the data license accordingly.

\section{Limitation and future work} \label{sm:limitation}
Although GeoSense represents a significant step forward in evaluating MLLMs on GPS tasks, there are several limitations that can be addressed for improvement. 

We have established a knowledge framework that includes 148 geometric principles covering most content from plane and solid geometry at the middle and high school levels. This framework allows us to measure MLLMs' ability to identify and apply basic geometric principles from multiple dimensions. However, future MLLMs are expected to master more complex knowledge across broader domains. Thus, extending GeoSense to include more challenging problems (such as those at the university level) is necessary for a more comprehensive and robust evaluation of MLLMs' multimodal reasoning abilities.

Additionally, we introduced two novel metrics, GPI and GPA, to assess models' mastery of geometric principles. However, the errors MLLMs make in GPA generally fall into two categories: insufficient perceptual ability with regard to chart elements, and incorrect mapping of geometric principles to geometric elements. Distinguishing between these two types of errors allows for a more granular understanding of MLLMs' reasoning capabilities from different perspectives.

\section{More Dataset Details } \label{sm:dataset}
\subsection{Geometric Principles Structure} \label{sm:cps}
Table \ref{tab:framework} illustrates the detailed framework of geometric principles within GeoSense, comprising 148 principles. Utilizing this framework and our annotations during problem-solving, we aim to evaluate the ability of MLLMs to accurately identify and apply geometric principles when tackling problems. Inspired by the human paradigm of problem-solving based on geometric principles, we construct our dataset using these principles as fundamental units, ensuring scientific rigor and valuable insights in the evaluation results.

\subsection{Examples with Geometric Principles Annotated in GeoSense} \label{sm:egpas}
\begin{figure*}[htbp]
    \centering
    \includegraphics[width=\linewidth]{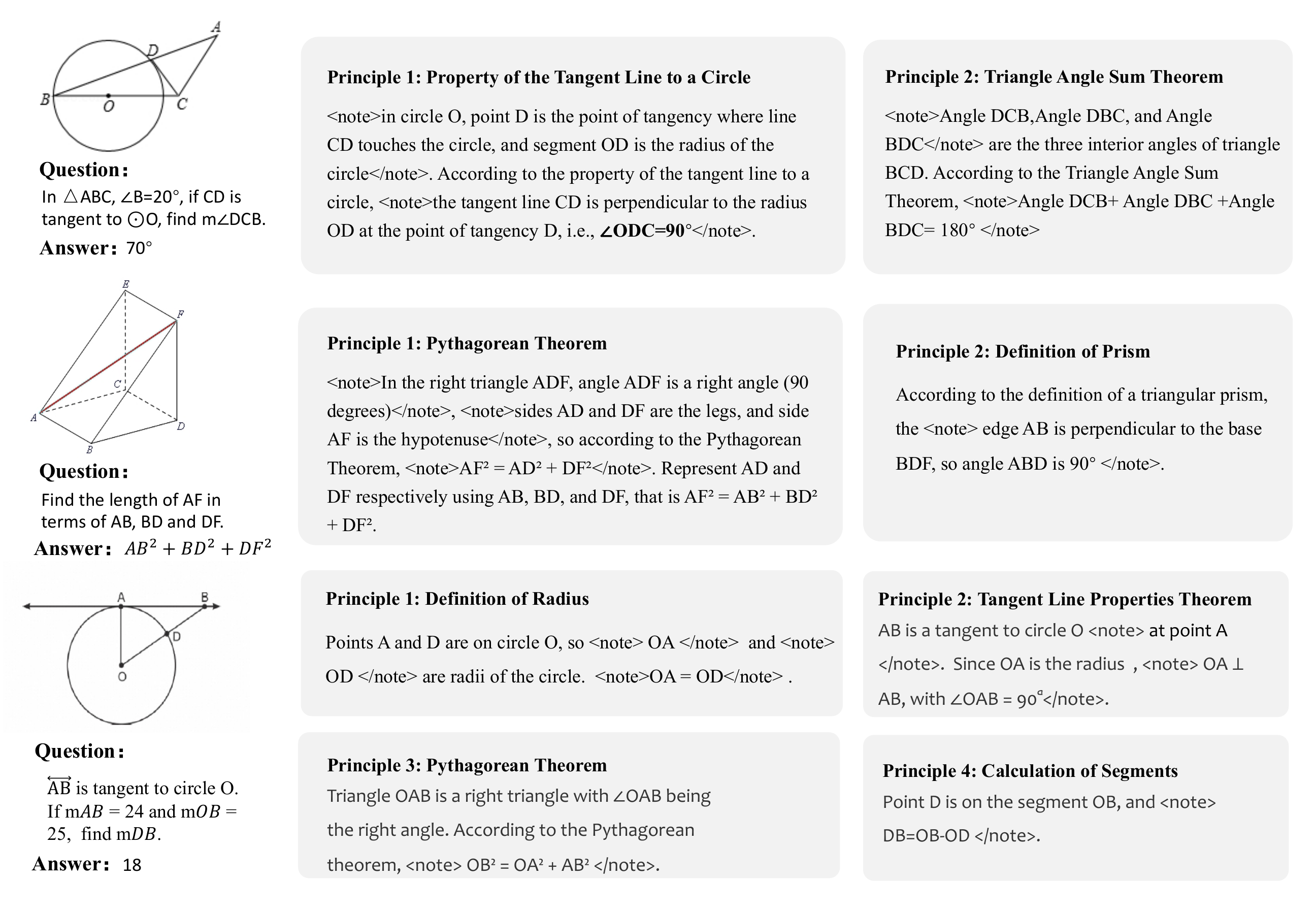}
    \Description{A woman and a girl in white dresses sit in an open car.}

  \caption{Example of GeoSense-English.}
  \label{fig:anno_sample}
\end{figure*}

\begin{figure*}[htbp]
    \centering
    \includegraphics[width=\linewidth]{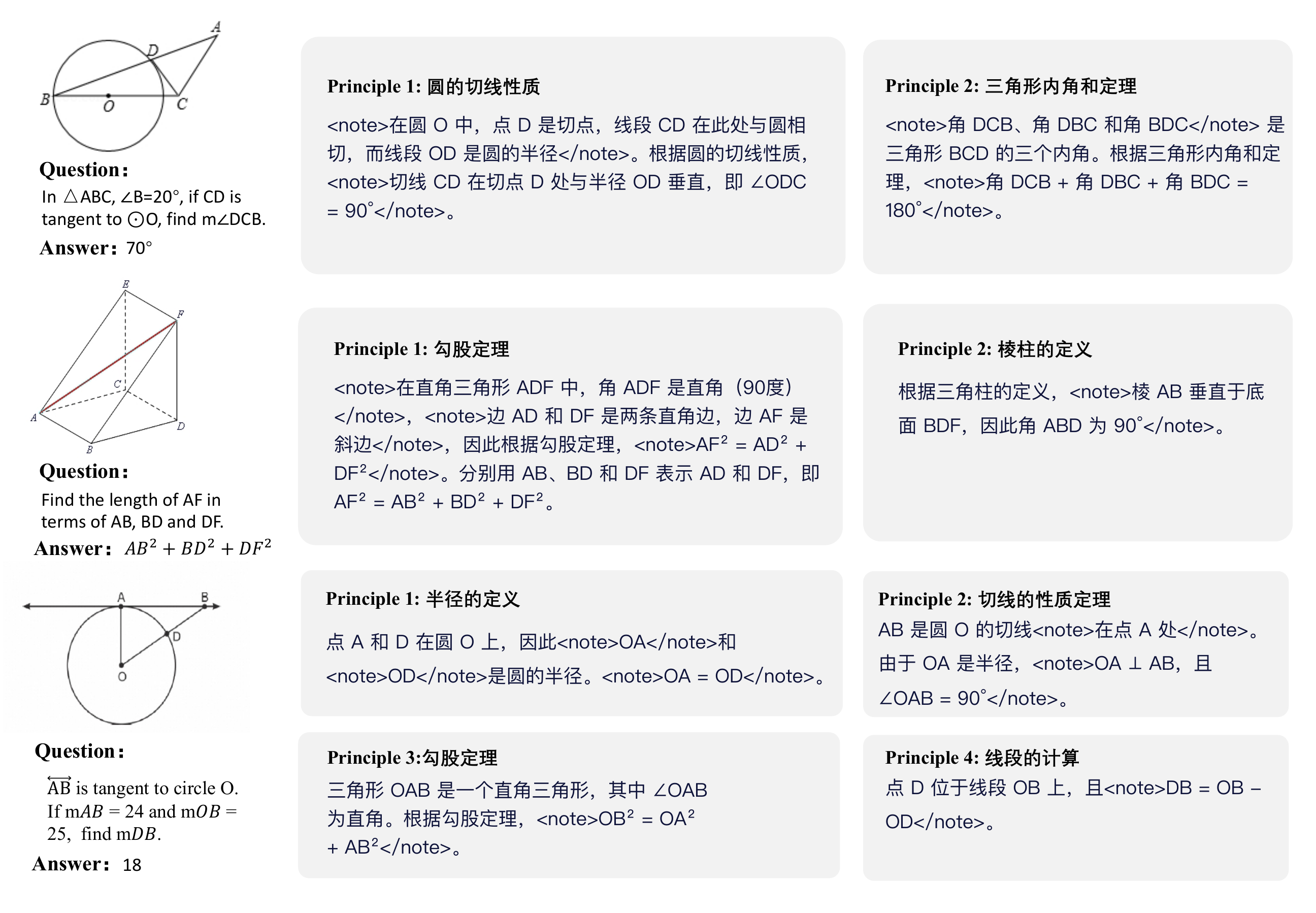}
    \Description{A woman and a girl in white dresses sit in an open car.}

  \caption{Example of GeoSense-Chinese.}
  \label{fig:anno_sample2}
\end{figure*}

Please refer to Figure \ref{fig:anno_sample} and Figure \ref{fig:anno_sample2}. We have annotated each geometry problem with the geometric principles needed for solving them, detailing the correspondence between these principles and elements within the diagram, along with their specific applications in the solution process. Notably, we support both Chinese and English versions.
\section{Additional experimental details} \label{sm:experimental}
\subsection{Evaluation Results of More MMLMs on GeoSense-English} \label{sm:ermge}
\begin{figure*}[htbp]
    \centering
    \includegraphics[width=\linewidth]{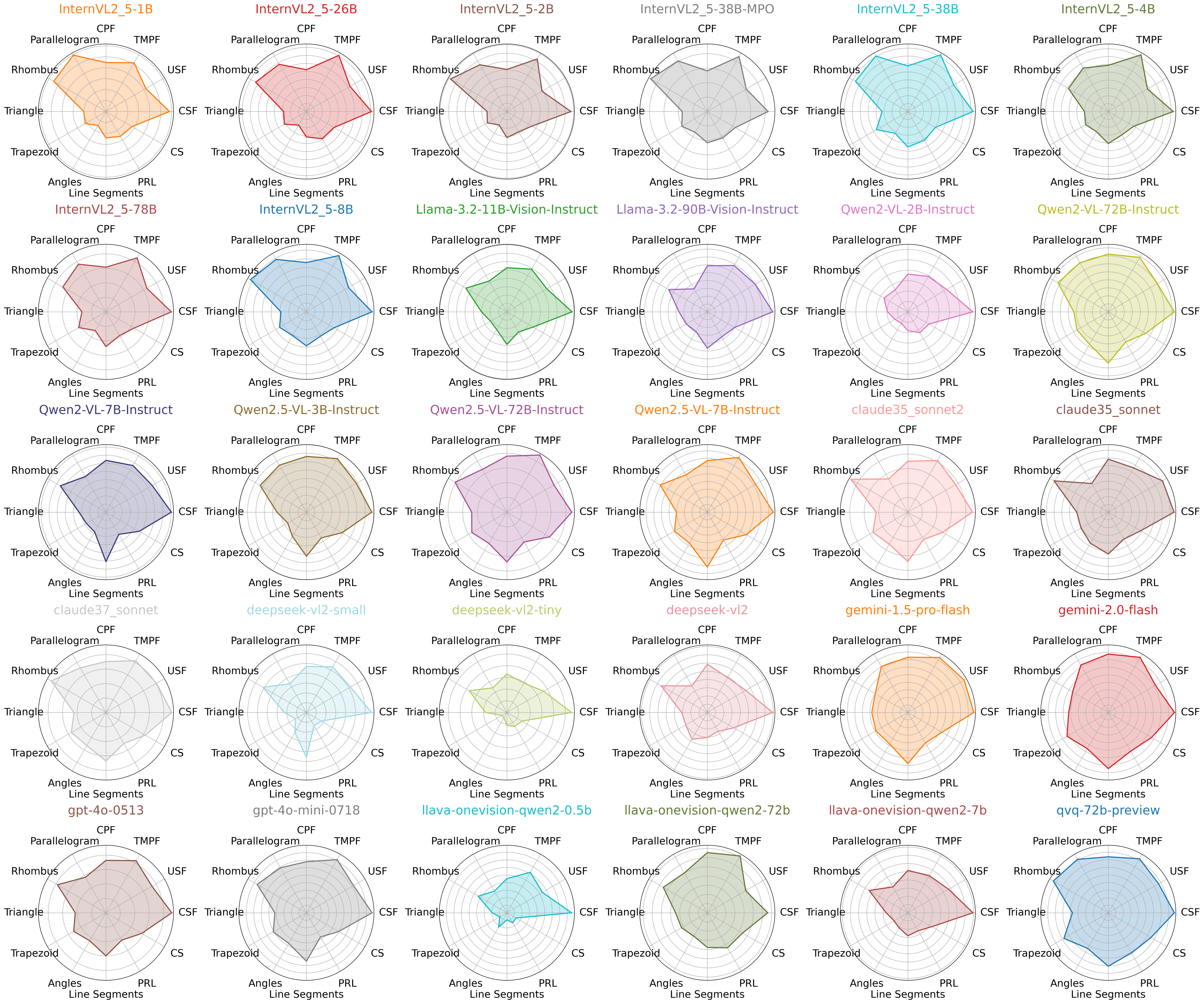}
    \Description{A woman and a girl in white dresses sit in an open car.}

  \caption{The performance of MLLMs in different subjects on GPI.}
  \label{fig:all_gpi_radra}
\end{figure*}

\begin{figure*}[htbp]
    \centering
    \includegraphics[width=\linewidth]{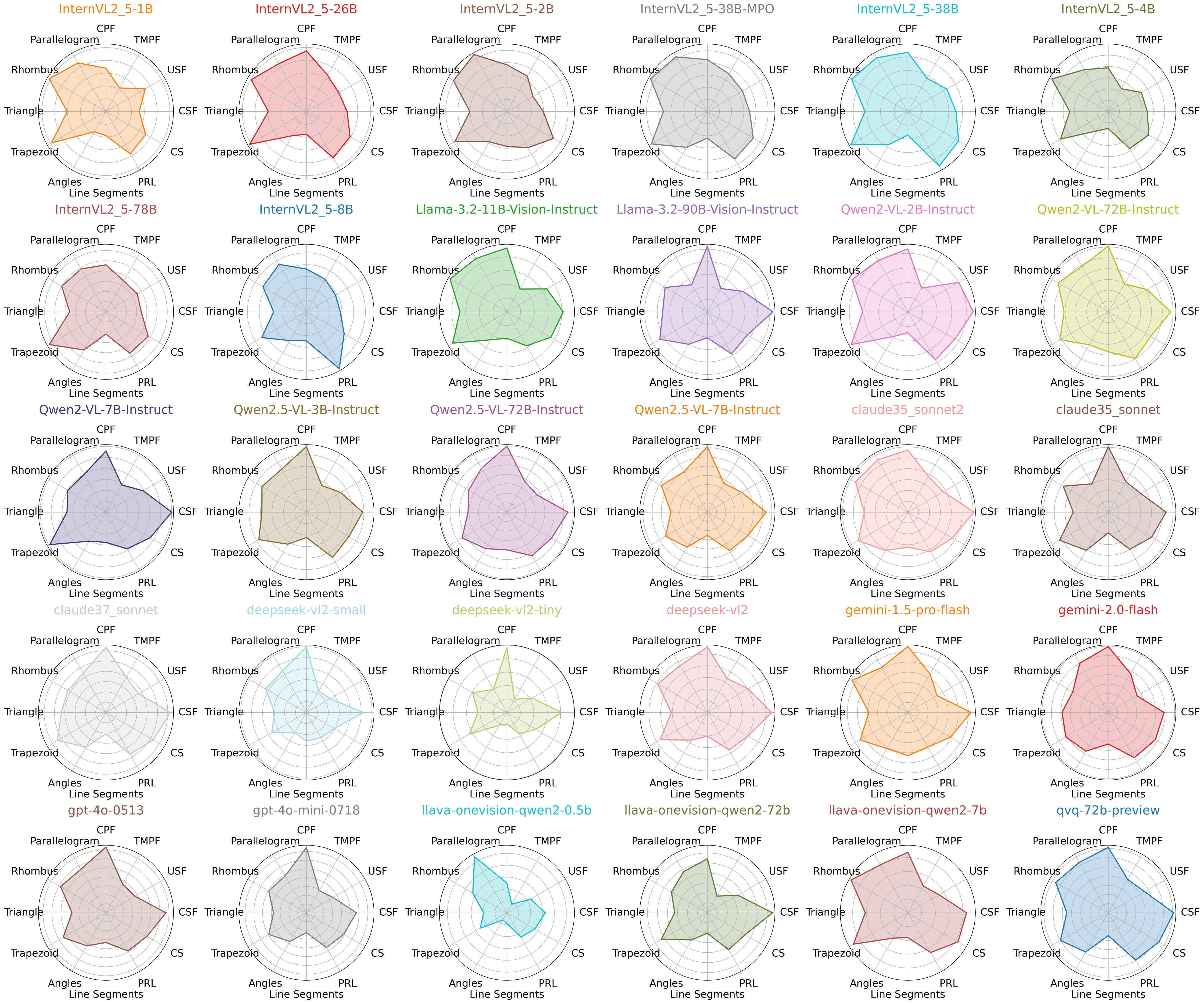}
    \Description{A woman and a girl in white dresses sit in an open car.}

  \caption{The performance of MLLMs in different subjects on GPA.}
  \label{fig:all_gpa_radra}
\end{figure*}

\begin{figure*}[htbp]
    \centering
    \includegraphics[width=\linewidth]{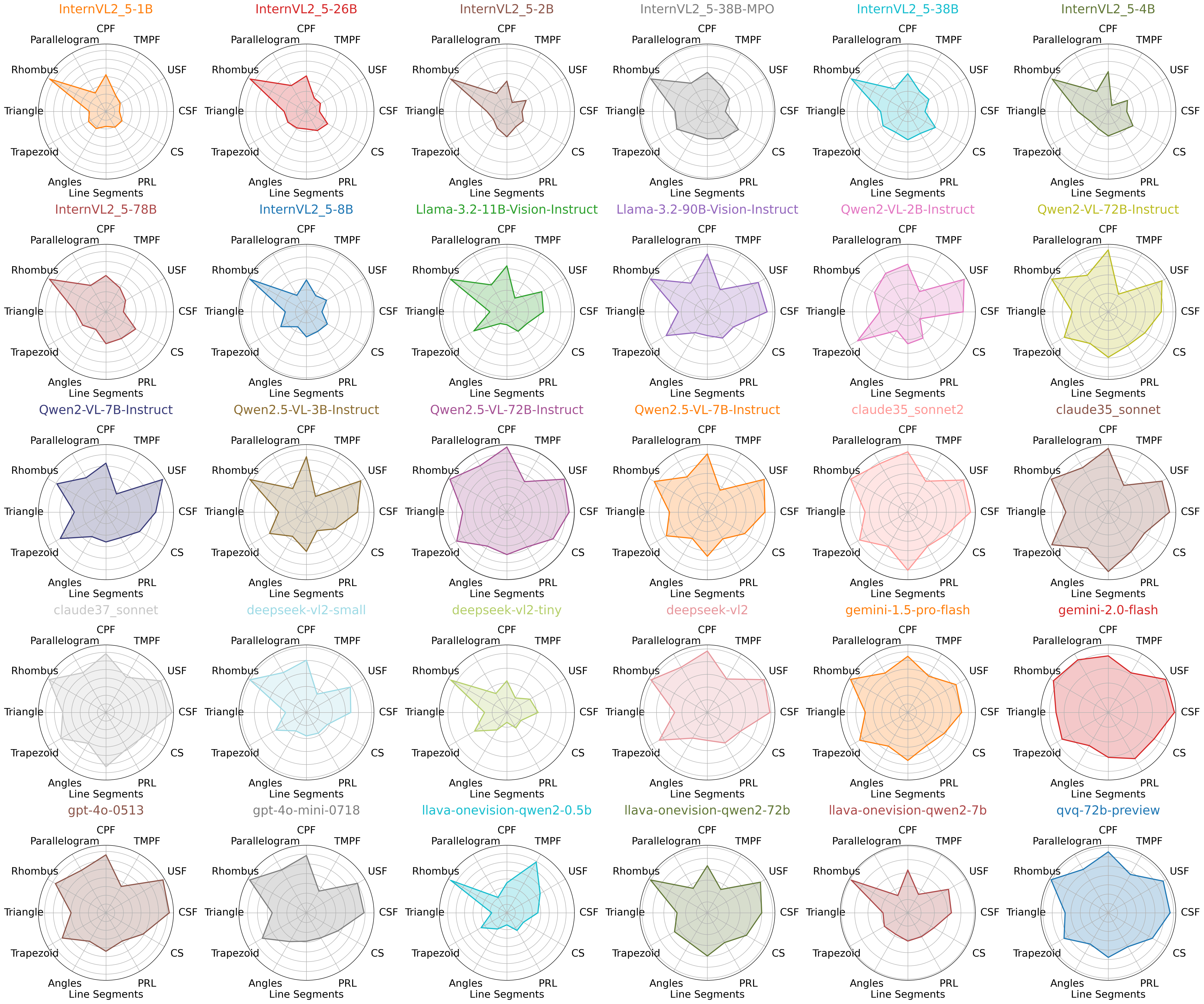}
    \Description{A woman and a girl in white dresses sit in an open car.}

  \caption{The performance of MLLMs in different subjects on ACC.}
  \label{fig:all_ans_radra}
\end{figure*}

Tables \ref{tab:en_all_results} and \ref{tab:en_all_results2} respectively show the performance of more MLLMs on GeoSenese-English under different geometric principle attributes and subjects. Additionally, we present the GPI, GPA, and ACC metrics of each MLLM across different subjects in Figures \ref{fig:all_gpi_radra}, \ref{fig:all_gpa_radra}, and \ref{fig:all_ans_radra}.
\subsection{Response of Differenet MLLMs} \label{sp:rdm}
\begin{figure*}[htbp]
    \centering
    \includegraphics[width=\linewidth]{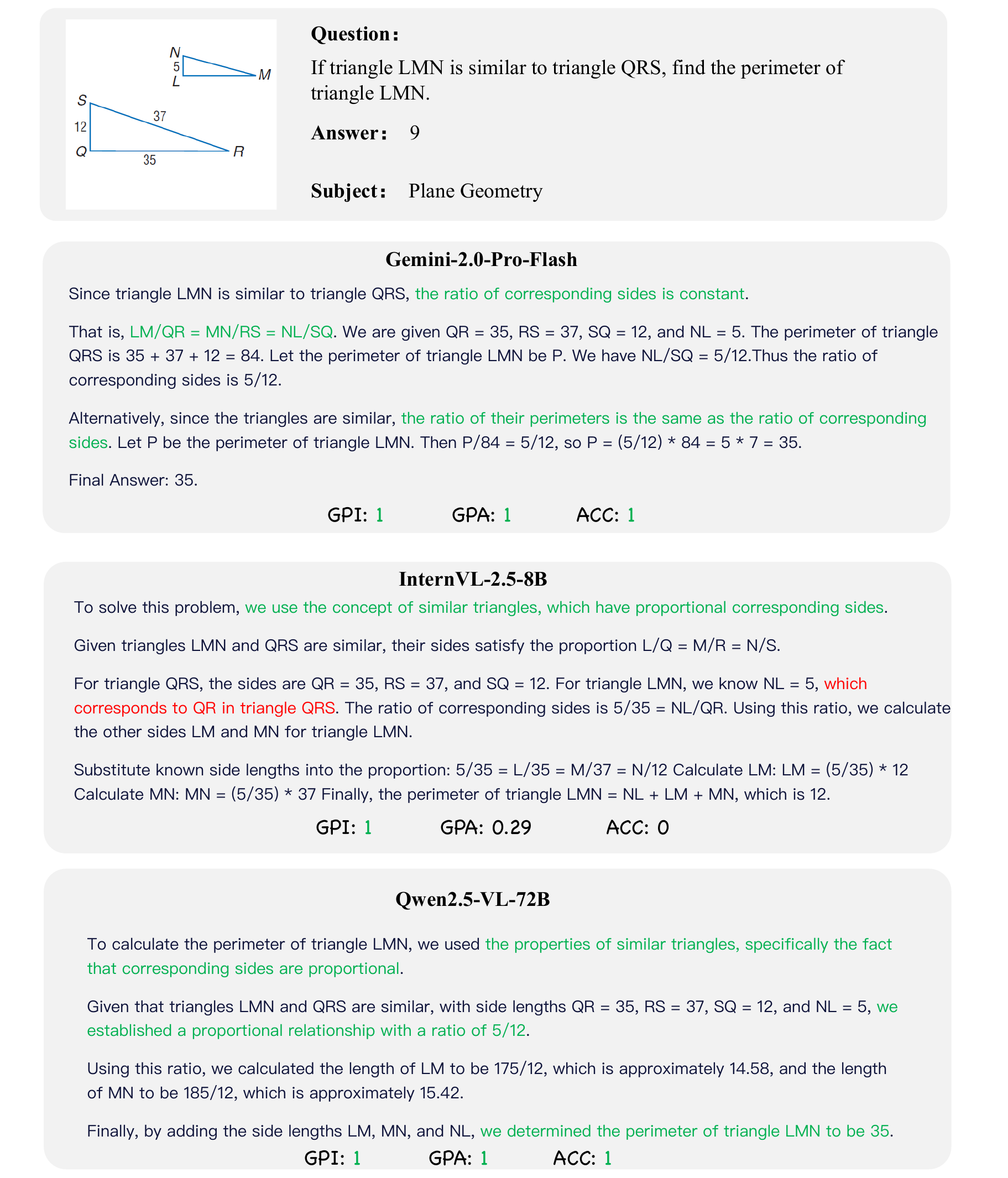}
    \Description{A woman and a girl in white dresses sit in an open car.}
  \caption{The performance of MLLMs in different subjects on GPI.}
  \label{fig:comcases}
\end{figure*}

Figure \ref{fig:comcases} shows the responses of different open-source and closed-source models to the same problem, along with their GPI, GPA, and ACC scores.
\subsection{Prompts for Evaluation Strategy} \label{sm:pes}
Table \ref{tab:eval_prompt} presents the prompts used in our evaluation process. These prompts guide the assessment of the model's ability to apply geometric principles accurately and comprehensively. The evaluation phases include verifying final answers, assessing geometric principle identification, extracting relevant content, and evaluating geometric principle alignment. This structured approach ensures precise and consistent evaluation of the model's proficiency in GPS.
\begin{tcolorbox}[colback=green!5!white, colframe=green!60!black, title=Evaluation Phases]
\textbf{Prompt for Evaluating Final Answers.} It involves verifying the correctness of the model's prediction against the provided correct answer. \\
\textbf{Prompt for Evaluating GPI.} It requires evaluators to rigorously assess whether the model has correctly utilized specific geometric principles within its predictive response. 

\textbf{Prompt for Extracting Relevant Content.} In this phase, evaluators focus on extracting all facets related to the geometric principles from the model's response. Ensuring that even dispersed but relevant content is captured, this prompt aids in discerning the comprehensiveness of the model's engagement with the principle. 

\textbf{Prompt for Evaluating GPA.} It delves into a deeper comparison between the model-generated description and the standard answer concerning specific geometric principles. The use of <note> </note> tags signifies key elements, helping evaluators measure the alignment in terms of presence and correctness. This prompt extends assessment to symbolic representations, which are critical in geometric evaluations.
\end{tcolorbox}
\onecolumn
\begin{longtable}{|>{\raggedright\arraybackslash}p{.15\linewidth}|>{\raggedright\arraybackslash}p{.15\linewidth}|>{\raggedright\arraybackslash}p{.15\linewidth}|>{\raggedright\arraybackslash}p{0.48\linewidth}|}
\hline
\rowcolor{gray!20}\textbf{Level 1} & \textbf{Level 2} & \textbf{Level 3} & \textbf{Level 4-5} \\ 
\hline
\endfirsthead

\hline
\rowcolor{gray!20}\textbf{Level 1} & \textbf{Level 2} & \textbf{Level 3} & \textbf{Level 4-5} \\ 
\hline
\endhead

\multirow{4}{*}{Solid Geometry} & Calculaiton of Solid Figures & Calculaiton of Surface Area of Solid Figures & 
\begin{itemize}[leftmargin=*]
\item Surface Area Formula for a Cube
\item Formula for the Surface Area of a Cone
\item Lateral Surface Area Formula of a Prism
\item Formula for Lateral Area of a Cylinder
\item Lateral Surface Area of a Cone
\item Surface Area Formula of a Prism
\item Surface Area Formula for Rectangular Prism
\item Surface Area Formula for a Cylinder
\item Sphere Surface Area Formula
\item Surface Area Formula for a Triangular Prism
\end{itemize} \\ 
\cline{3-4}
 &  & Calculaiton of Volume of Solid Figures & 
\begin{itemize}[leftmargin=*]
\item Volume Formula of Rectangular Prism
\item Volume Formula of a Cube
\item Volume Formula of Prism
\item Volume Formula of Pyramid
\item Volume Formula of Cylinder
\item Volume Formula of a Cone
\item Formula for the Volume of a Sphere
\end{itemize} \\ 
\cline{2-4}
 & Understanding of Solid Figures & Cylinder and Cone & 
\begin{itemize}[leftmargin=*]
\item Definition of Cone
\item Generatrix
\item Cylinder
\item Development of a Cone
\end{itemize} \\ 
\cline{3-4}
 &  & Rectangular Prism and Cube & 
\begin{itemize}[leftmargin=*]
\item Definition of Rectangular Prism
\item Definition of Cube
\end{itemize} \\ 
\cline{3-4}
 &  & Prism & 
\begin{itemize}[leftmargin=*]
\item Definition of Prism
\item Definition of Triangular Prism
\end{itemize} \\ 
\cline{3-4}
 &  & Sphere & 
\begin{itemize}[leftmargin=*]
\item Definition of Sphere
\item Radius of a Sphere
\end{itemize} \\ 
\hline
\multirow{8}{*}{Plane Geometry} & Transformation and Motion of Plane Figures & Symmetry & 
\begin{itemize}[leftmargin=*]
\item Symmetric Point
\item Law of Reflection
\item Definition of Rotation
\item Properties of Rotation
\item Definition of Scale Factor in Rectangular Coordinate Systems
\item Rotation Invariance Theorem
\item Definition of Rotational Symmetry
\item Rotation Transformation
\item 2D Plane Rotation Formula
\item Reflection Transformation
\item Definition of Translation
\item Translation Invariance Theorem
\item Scaling Theorem in a Rectangular Coordinate System
\end{itemize} \\ 
\cline{2-4}
 & Calculation of Plane Figures & Interior and Exterior Angles of Polygon & 
\begin{itemize}[leftmargin=*]
\item Triangle Angle Sum Theorem
\item Exterior Angle Sum Theorem of Polygon
\item Formulas for the Central Angle and Interior Angle of a Regular Polygon
\item Sum of Interior Angles of a Quadrilateral Theorem
\item Polygon Interior Angle Sum Theorem
\end{itemize} \\ 
\cline{3-4}
 &  & Calculation of Areas & 
\begin{itemize}[leftmargin=*]
\item Formula for the Area of a Rectangle
\item Triangle Area Formula (Using Sine Function)
\item Area Formula of a Circle
\item Area Formula of a Triangle
\item Theorem on the Area Ratio of Similar Triangles
\item Area Formula of a Parallelogram
\item Trapezoid Area Formula
\item Rhombus Area Formula
\item Heron's Formula
\item Area Formula for Square
\end{itemize} \\ 
\cline{3-4}
 &  & Calculation of Perimeters & 
\begin{itemize}[leftmargin=*]
\item Formula for the Perimeter of a Rectangle
\item Circumference Formula of Circle
\item Formula for the Length of an Arc of a Sector
\item Formula for the Area of a Sector
\item Perimeter of a Parallelogram
\item Arc Length Formula of a Circle
\end{itemize} \\ 
\cline{2-4}
 & Understanding of Plane Figures & Polygons & 
\begin{itemize}[leftmargin=*]
\item Definition of Regular Polygon
\end{itemize} \\ 
\cline{3-4}
 &  & Properties and Understanding of Parallelogram & 
\begin{itemize}[leftmargin=*]
\item Definition of Parallelogram
\item Properties of Parallelogram Theorem
\item Adjacent Angles Supplementary Theorem of Parallelogram
\end{itemize} \\ 
\cline{3-4}
 &  & Properties and Understanding of Rhombus & 
\begin{itemize}[leftmargin=*]
\item Definition of Rhombus
\item Properties of the Diagonals of a Rhombus
\end{itemize} \\ 
\cline{3-4}
 &  & Properties and Understanding of Triangle & 
\begin{itemize}[leftmargin=*]
\item General Properties of Triangle
\begin{itemize}
\item Definition of Triangle
\item Definition of Equilateral Triangle
\item Definition of Median of a Triangle
\item Triangle Midline Theorem
\item Pythagorean Theorem
\item Exterior Angle Theorem of Triangle
\item Definition of Right Triangle
\end{itemize}
\item Trigonometric Functions
\begin{itemize}
\item Sine Theorem
\item Definition of Sine Function
\item Definition of Tangent Function
\item Cosine Function
\item Properties of 30°-60°-90° Triangle
\item Complementary Acute Angles in a Right Triangle
\end{itemize}
\item Properties and Understanding of Isosceles Triangle
\begin{itemize}
\item Definition of Isosceles Triangle
\item Properties of Isosceles Triangle
\item Coincidence Theorem of Altitude, Median, and Angle Bisector in Isosceles Triangle
\item Definition of Isosceles Right Triangle
\end{itemize}
\item Properties and Understanding of Similar Triangles
\begin{itemize}
\item Definition of Similar Triangles
\item Similarity Theorem for Triangles (AA)
\item SAS Criterion for Similar Triangles
\item Similarity Theorem for Triangles (SSS)
\end{itemize}
\item Properties and Understanding of Congruent Triangles
\begin{itemize}
\item Definition of Congruent Triangles
\item Triangle Congruence Theorem (SSS)
\item Triangular Congruence Theorem (SAS)
\item Congruence Theorem for Triangles (AAS)
\item Angle-Side-Angle (ASA) Criterion for Congruence of Triangle
\item Right Triangle Congruence (Hypotenuse, One Leg)
\end{itemize}
\end{itemize} \\ 
\cline{3-4}
 &  & Properties and Understanding of Trapezoid & 
\begin{itemize}[leftmargin=*]
\item Definition of Trapezoid
\item Median Line Theorem of Trapezoid
\item Definition of Isosceles Trapezoid
\item Properties of an Isosceles Trapezoid
\end{itemize} \\ 
\cline{3-4}
 &  & Properties and Understanding of Rectangle & 
\begin{itemize}[leftmargin=*]
\item Definition of Rectangle
\item Property of Diagonals in a Rectangle
\item Definition of Square
\end{itemize} \\ 
\cline{2-4}
 & Representation and Understanding of Angles & Representation and Understanding of Angles & 
\begin{itemize}[leftmargin=*]
\item Definition of Vertical Angles
\item Definition of Straight Angle
\item Definition of Linear Pair of Angles
\item Definition of Exterior Angle of a Polygon
\item Definition of Angle Bisector
\item Naming of Angles
\item Measurement of Angle
\end{itemize} \\ 
\cline{2-4}
 & Properties and Understanding of Line Segments & Midpoint of a Line Segment & 
\begin{itemize}[leftmargin=*]
\item Definition of Line Segment
\item Definition of Ray
\item Distance Formula Between Two Points
\end{itemize} \\ 
\cline{2-4}
 & Positional Relationships between Lines & Perpendicularity & 
\begin{itemize}[leftmargin=*]
\item Definition of Perpendicular Lines
\item Definition of Altitude
\item Definition of Perpendicular Bisector
\item Properties of the Perpendicular Bisector
\item Definition of Foot of a Perpendicular
\end{itemize} \\ 
\cline{3-4}
 &  & Parallel & 
\begin{itemize}[leftmargin=*]
\item Definition of Parallel Lines
\item Parallel Postulate 2 of Parallel Lines
\item Definition of Corresponding Angles
\item Definition of Alternate Interior Angles
\item Proportional Segments Theorem
\item Definition of Consecutive Interior Angles
\item Transitivity of Parallel Lines
\end{itemize} \\ 
\cline{2-4}
 & Circles and Sectors &Properties and Understanding of Circles and Sectors & 
\begin{itemize}[leftmargin=*]
\item Definition of Circle
\item Definition of Radius
\item Definition of Diameter
\item Corollary to the Inscribed Angle Theorem 2: The Angle Subtended by the Diameter
\item Definition of Semicircle
\item Definition of Central Angle
\item Definition of Inscribed Angle
\item Inscribed Angle Theorem
\item Definition of Tangent to a Circle
\item Tangent-Segment Theorem
\item Cyclic Quadrilateral
\item Corollary 3 of the Inscribed Angle Theorem: Diagonal Supplementary Theorem for Cyclic Quadrilateral
\item Property of the Tangent Line to a Circle
\item Perpendicular Diameter Theorem
\item Corollary 1 of the Inscribed Angle Theorem
\item Definition of Chord
\item Definition of Arc
\item Central Angle Theorem
\item Properties of Central Angles
\item Secant Line Theorem
\item Formula for Conversion between Degrees and Radians
\end{itemize} \\ 
\hline
\caption{Hierarchical Structure of Geometric Principles}
\label{tab:framework}
\end{longtable}
\twocolumn

\begin{table*}[htbp]
\centering
\begin{tabular}{p{4cm}|p{12cm}}
\hline
\textbf{Phase} & \textbf{Prompt} \\
\hline
Evaluate Final Answers & You are a geometry assessment expert. Given the model's prediction and the correct answer, determine whether the model's predicted answer is correct. 

Model Response: \{predict\}  \quad Answer: \{answer\}

Please provide a clear response with 'yes' or 'no'.
\\
\hline
Evaluate GPI & 
Please rigorously assess whether the model's predicted response utilizes the specified geometric principle:

Principle Name: \{name\} \quad Principle Content: \{content\} \quad Model Response: \{predict\}

Please respond clearly with 'yes' or 'no'. \\
\hline
Extract relevant content & Based on the following geometric principle, extract all related content from the model's response (extract any related content even if it is not contiguous):

Principle Name: \{name\} \quad Principle Content: \{content\} \quad Model Response: \{predict\}

Please directly output all content from the model's response that is related to this principle. \\
\hline
Evaluate GPA & You are a geometry assessment expert. Your task is to evaluate the model-generated description against the standard answer for a specific geometric principle. Specifically, in the standard answer, the content wrapped in <note> </note> represents key elements. You need to assess whether the model’s response includes these key elements and, if so, whether they correspond correctly—paying particular attention to symbolic representation (e.g., angles ABC and CBA refer to the same angle, which should not be marked incorrect due to order).

Finally, please output [ans]num\_exist, num\_acc, ground\_total[/ans], where num\_exist is the number of instances of these key elements (as indicated by the tags) present in the model response (regardless of correctness), num\_acc is the count of instances that are present and correctly correspond, and ground\_total is the total number of key elements wrapped in <note> </note> tags in the standard answer.

Standard Description: \{ground\_truth\}

Relevant Steps in Model Response: \{response\} \\
\hline
\end{tabular}
\caption{Prompts for Evaluation Strategy.}
\label{tab:eval_prompt}
\end{table*}

\begin{table*}[htbp]
  \centering
    \begin{tabular}{l|ccc|ccc|ccc|cccc}
    \toprule
    \multicolumn{1}{c|}{\multirow{2}[4]{*}{\textbf{Model}}} & \multicolumn{3}{c|}{\textbf{Definitions}} & \multicolumn{3}{c|}{\textbf{Theorems}} & \multicolumn{3}{c|}{\textbf{Formula}} & \multicolumn{4}{c}{\textbf{ALL}} \\
\cmidrule{2-14}          & GPI   & GPA   & ACC   & GPI   & GPA   & ACC   & GPI   & GPA   & ACC   & GPI   & GPA   & ACC   & AVG \\
    \midrule
    \multicolumn{14}{c}{\textbf{Closed-Sourced MLLMs}} \\
    \midrule
    Claude35\_Sonnet & 56.5  & 41.2  & 41.9  & 54.9  & 46.8  & 33.8  & 82.8  & 52,5  & 52.9  & 63.2  & 40.8  & 46.1  & 50.0  \\
    Claude35\_Sonnet2 & 58.8  & 44.0  & 49.6  & 59.0  & 46.0  & 42.5  & 84.7  & 52.7  & 59.2  & 66.3  & 43.1  & 53.7  & 54.4  \\
    Claude37\_Sonnet & 62.0  & 46.7  & 54.3  & 60.2  & 50.0  & 46.5  & {92.4 } & 56.1  & 67.9  & 68.7  & 45.2  & 57.6  & 57.2  \\
    Gemini-1.5-pro-flash & 60.2  & 43.8  & 53.0  & 58.7  & 51.5  & 45.6  & 85.9  & 55.3  & 56.1  & 67.9  & 44.9  & 55.7  & 56.2  \\
    Gemini-2.0-pro-flash & 64.2  & 47.0  & {73.3 } & {72.7 } & {59.0 } & {72.4 } & 87.4  & {60.0 } & {77.9 } & 72.1  & 49.7  & {74.1 } & {65.3 } \\
    GPT-4o & 56.3  & 46.3  & 48.0  & 54.1  & 49.3  & 37.4  & 90.8  & 58.3  & 61.1  & 64.4  & 45.3  & 51.7  & 53.8  \\
    GPT-4o-mimi & 54.1  & 44.1  & 40.9  & 48.6  & 49.9  & 34.0  & 86.6  & 50.1  & 49.6  & 61.3  & 41.7  & 43.7  & 48.9  \\
    \midrule
    \multicolumn{14}{c}{\textbf{Open-Soured MLLMs}} \\
    \midrule
    InterVL2.5-1B & 31.3  & 37.1  & 11.1  & 28.0  & 36.4  & 9.6   & 58.8  & 24.3  & 6.9   & 34.5  & 21.5  & 11.1  & 22.4  \\
    InterVL2.5-2B & 39.2  & 42.7  & 15.5  & 32.4  & 42.5  & 14.3  & 67.9  & 30.1  & 11.4  & 42.2  & 27.5  & 15.9  & 28.5  \\
    InterVL2.5-8B & 43.7  & 40.7  & 21.2  & 38.2  & 39.3  & 20.0  & 67.6  & 24.7  & 13.7  & 46.7  & 27.3  & 21.1  & 31.7  \\
    InterVL2.5-26B & 44.1  & 40.1  & 23.8  & 39.2  & 39.9  & 25.1  & 74.8  & 25.1  & 11.1  & 47.2  & 28.1  & 22.6  & 32.6  \\
    InterVL2.5-38B & 48.7  & 40.6  & 28.9  & 44.5  & 43.9  & 29.8  & 74.8  & 26.4  & 16.0  & 52.7  & 31.1  & 27.3  & 37.0  \\
    InterVL2.5-38B-MPO$\dagger$ & 50.7  & 44.6  & 29.7  & 48.2  & 46.4  & 30.0  & 75.6  & 29.3  & 16.0  & 53.9  & 33.6  & 27.7  & 38.4  \\
    InterVL2.5-78B & 49.0  & 45.2  & 29.8  & 48.6  & 46.8  & 32.0  & 80.2  & 30.5  & 18.3  & 53.7  & 32.9  & 28.7  & 38.4  \\
\cmidrule{1-13}    Deepseek-VL2-tiny & 20.1  & 27.4  & 15.8  & 20.1  & 27.4  & 15.8  & 53.4  & 35.9  & 18.7  & 17.6  & 26.5  & 15.2  & 19.8  \\
    Deepseek-VL2-small & 25.6  & 35.7  & 23.3  & 26.7  & 36.1  & 19.5  & 67.9  & 48.1  & 30.2  & 34.2  & 23.8  & 26.3  & 28.1  \\
    Deepseek-VL2 & 40.1  & 37.8  & 33.1  & 40.6  & 39.6  & 26.0  & 76.3  & 52.8  & 42.4  & 48.4  & 33.4  & 35.7  & 39.2  \\
\cmidrule{1-13}    Llama-vision-11B & 43.2  & 36.1  & 22.6  & 37.9  & 35.6  & 18.7  & 74.8  & 37.5  & 29.8  & 47.9  & 29.2  & 24.8  & 34.0  \\
    Llama-vision-90B & 49.1  & 39.2  & 27.3  & 42.0  & 36.0  & 21.2  & 78.2  & 43.6  & 37.0  & 52.9  & 31.4  & 29.8  & 38.0  \\
\cmidrule{1-13}    LLaVA-onevison-0.6B & 11.8  & 32.6  & 37.8  & 10.5  & 26.6  & 5.0   & 43.1  & 30.9  & 7.6   & 17.1  & 10.1  & 6.3   & 11.2  \\
    LLaVA-onevison-7B & 36.3  & 38.0  & 22.7  & 39.2  & 39.2  & 22.7  & 72.9  & 40.6  & 42.6  & 41.4  & 26.0  & 22.8  & 30.1  \\
    LLaVA-onevison-72B & 47.9  & 39.0  & 33.7  & 49.6  & 44.8  & 36.4  & 68.3  & 55.9  & 43.1  & 52.5  & 33.2  & 37.2  & 41.0  \\
    \cmidrule{1-13}    Qwen2-VL-2B & 28.1  & 36.3  & 10.9  & 25.1  & 39.0  & 8.4   & 60.3  & 44.6  & 16.0  & 32.5  & 22.6  & 12.5  & 22.5  \\
    Qwen2-VL-7B & 48.2  & 41.7  & 31.9  & 48.0  & 43.2  & 27.5  & 81.3  & 55.2  & 38.2  & 54.6  & 36.8  & 33.6  & 41.7  \\
    Qwen2-VL-72B & 57.2  & 44.2  & 46.6  & 57.7  & 44.2  & 46.6  & 85.5  & 52.0  & 50.4  & 64.0  & 43.4  & 49.2  & 52.2  \\
    Qwen2.5-VL-3B & 50.5  & 39.9  & 33.5  & 48.8  & 47.0  & 27.7  & 74.8  & 45.0  & 41.2  & 55.2  & 36.5  & 34.9  & 42.2  \\
    Qwen2.5-VL-7B & 57.7  & 45.6  & 43.6  & 57.4  & 51.2  & 37.5  & 85.9  & 60.4  & 53.1  & 63.1  & 44.6  & 46.3  & 51.3  \\
    Qwen2.5-VL-72B & 61.5  & 47.5  & 61.5  & 65.1  & 54.8  & 57.5  & 89.7  & 61.5  & 63.8  & 68.5  & 48.1  & 63.8  & 60.1  \\
    QVQ-72B-Preview$\dagger$ & {68.2 } & {56.0 } & 53.1  & 63.6  & 58.3  & 49.6  & 85.1  & 58.4  & 54.2  & {72.3 } & {53.5} & 54.3  & 60.0  \\
    \bottomrule
    \end{tabular}%
    \caption{Evaluation Results on Different Types of Geometric Principles of More MMLMs on GeoSense-English. GPI = Geometric Principles Identification, GPA= Geometric Principles Application. MLLMs with $\dagger$ are typically trained for reasoning tasks.}
  \label{tab:en_all_results}%
\end{table*}%
\begin{table*}[htbp]
  \centering
    \begin{tabular}{l|ccc|ccc|ccc|ccc|ccc}
    \toprule
    \multicolumn{1}{c|}{\multirow{2}[4]{*}{\textbf{Model}}} & \multicolumn{3}{c|}{\textbf{CSF}} & \multicolumn{3}{c|}{\textbf{USF}} & \multicolumn{3}{c|}{\textbf{TMPF}} & \multicolumn{3}{c|}{\textbf{CPF}} & \multicolumn{3}{c}{\textbf{UPF}} \\
\cmidrule{2-16}          & K.I.  & K.A.  & ACC.  & K.I.  & K.A.  & ACC.  & K.I.  & K.A.  & ACC.  & K.I.  & K.A.  & ACC   & K.I.  & K.A.  & ACC \\
    \midrule
    \multicolumn{16}{c}{\textbf{Closed-Sourced MLLMs}} \\
    \midrule
    Claude35\_Sonnet & 85.0  & 53.5  & 53.8  & 80.9  & 34.7  & 54.6  & 65.9  & 32.5  & 27.2  & 68.4  & 60.5  & 56.1  & 45.1  & 38.7  & 40.7  \\
    Claude35\_Sonnet2 & 89.1  & 60.7  & 65.2  & 76.0  & 37.6  & 66.9  & 82.2  & 37.9  & 37.1  & 70.0  & 57.0  & 62.8  & 50.3  & 43.4  & 46.7  \\
    Claude37\_Sonnet & 91.1  & 62.5  & 76.8  & 80.8  & 36.5  & 73.5  & 82.7  & 39.4  & 47.3  & 70.5  & 63.5  & 68.5  & 54.1  & 44.8  & 52.0  \\
    Gemini-1.5-pro-flash & 87.3  & 62.6  & 62.1  & 86.0  & 33.5  & 64.2  & 83.9  & 44.1  & 48.4  & 73.1  & 65.4  & 64.8  & 50.7  & 43.5  & 49.3  \\
    Gemini-2.0-pro-flash & 88.1  & 58.8  & 89.9  & 72.4  & 35.1  & 89.9  & 84.6  & 47.2  & 62.1  & 77.7  & 69.5  & 77.0  & 60.3  & 51.2  & 70.9  \\
    GPT-4o & 91.3  & 66.8  & 72.3  & 73.7  & 37.0  & 75.1  & 83.5  & 37.0  & 34.7  & 72.7  & 73.1  & 66.1  & 49.4  & 44.8  & 44.2  \\
    GPT-4o-mimi & 87.3  & 56.9  & 58.8  & 72.3  & 35.0  & 60.3  & 81.7  & 29.2  & 25.6  & 67.9  & 74.4  & 58.5  & 46.6  & 41.6  & 36.9  \\
    \midrule
    \multicolumn{16}{c}{\textbf{Open-Soured MLLMs}} \\
    \midrule
    InterVL2.5-1B & 58.1  & 26.0  & 7.7   & 42.1  & 35.5  & 9.4   & 51.5  & 21.3  & 11.0  & 44.8  & 33.9  & 21.2  & 25.6  & 32.9  & 11.6  \\
    InterVL2.5-2B & 77.0  & 29.7  & 11.9  & 48.7  & 24.6  & 17.5  & 72.9  & 33.4  & 8.3   & 50.5  & 38.2  & 24.1  & 30.4  & 36.1  & 16.6  \\
    InterVL2.5-8B & 74.3  & 28.7  & 14.2  & 55.0  & 28.5  & 21.2  & 73.6  & 31.9  & 17.0  & 55.9  & 36.3  & 29.1  & 35.0  & 35.3  & 21.5  \\
    InterVL2.5-26B & 81.2  & 31.7  & 13.0  & 62.8  & 29.1  & 16.0  & 80.9  & 32.8  & 15.4  & 52.2  & 47.3  & 35.4  & 35.1  & 34.2  & 23.5  \\
    InterVL2.5-38B & 82.4  & 35.3  & 17.1  & 67.0  & 32.9  & 24.1  & 82.7  & 28.0  & 23.2  & 57.5  & 43.3  & 37.4  & 39.8  & 36.7  & 28.6  \\
    InterVL2.5-38B-MPO$\dagger$ & 84.0  & 34.8  & 16.4  & 62.1  & 33.5  & 23.3  & 87.9  & 35.7  & 26.1  & 56.3  & 42.8  & 35.6  & 41.9  & 39.7  & 30.1  \\
    InterVL2.5-78B & 90.1  & 34.5  & 17.4  & 65.0  & 35.4  & 22.5  & 86.0  & 34.8  & 27.6  & 61.5  & 46.0  & 36.2  & 40.2  & 41.7  & 30.8  \\
    \midrule
    Deepseek-VL2-tiny & 55.3  & 40.7  & 21.2  & 36.4  & 21.5  & 18.0  & 29.1  & 11.3  & 11.4  & 32.9  & 48.3  & 21.5  & 15.4  & 20.8  & 14.5  \\
    Deepseek-VL2-small & 66.3  & 51.8  & 34.1  & 52.0  & 25.3  & 38.9  & 53.7  & 22.3  & 16.3  & 47.0  & 59.7  & 40.1  & 21.9  & 28.8  & 20.0  \\
    Deepseek-VL2 & 79.4  & 55.0  & 49.2  & 53.7  & 40.3  & 51.7  & 49.0  & 33.6  & 30.4  & 58.6  & 56.0  & 48.3  & 33.6  & 35.6  & 28.6  \\
    \midrule
    Llama-vision-11B & 77.9  & 41.0  & 33.9  & 55.0  & 33.4  & 37.3  & 58.9  & 19.1  & 14.8  & 52.6  & 46.4  & 42.7  & 32.2  & 33.7  & 20.7  \\
    Llama-vision-90B & 83.4  & 52.3  & 45.8  & 70.5  & 32.7  & 45.1  & 68.2  & 21.5  & 19.7  & 58.8  & 52.3  & 44.3  & 37.5  & 34.8  & 24.1  \\
    \midrule
    LLaVA-onevison-0.6B & 42.8  & 32.3  & 7.3   & 27.1  & 23.0  & 9.1   & 31.0  & 8.3   & 13.9  & 22.6  & 25.0  & 7.1   & 9.3   & 22.4  & 5.6  \\
    LLaVA-onevison-7B & 79.3  & 42.6  & 32.8  & 57.3  & 26.2  & 35.3  & 52.3  & 22.4  & 16.0  & 51.6  & 43.8  & 32.2  & 28.6  & 33.6  & 21.7  \\
    LLaVA-onevison-72B & 65.7  & 65.5  & 50.0  & 48.1  & 35.3  & 56.3  & 71.4  & 19.3  & 24.8  & 65.5  & 54.2  & 43.3  & 39.1  & 37.7  & 35.0  \\
    \midrule
    Qwen2-VL-2B & 62.5  & 44.6  & 20.0  & 43.1  & 40.3  & 23.5  & 39.4  & 18.7  & 8.5   & 36.3  & 43.1  & 17.3  & 20.1  & 32.9  & 9.9  \\
    Qwen2-VL-7B & 80.5  & 60.0  & 43.2  & 65.7  & 39.2  & 56.7  & 66.2  & 28.7  & 18.3  & 63.7  & 55.9  & 42.6  & 38.6  & 38.7  & 31.0  \\
    Qwen2-VL-72B & 82.5  & 57.8  & 57.4  & 70.7  & 41.2  & 67.1  & 78.6  & 29.8  & 22.2  & 72.1  & 61.0  & 67.1  & 49.1  & 43.2  & 43.7  \\
    Qwen2.5-VL-3B & 77.6  & 48.9  & 48.7  & 68.4  & 34.3  & 60.0  & 73.4  & 26.8  & 17.4  & 66.1  & 57.1  & 53.1  & 40.6  & 39.5  & 29.5  \\
    Qwen2.5-VL-7B & 85.9  & 63.9  & 59.6  & 72.9  & 43.1  & 67.9  & 82.3  & 36.0  & 26.3  & 67.2  & 71.1  & 60.3  & 48.9  & 44.5  & 40.6  \\
    Qwen2.5-VL-72B & 88.2  & 64.4  & 74.1  & 73.9  & 36.6  & 78.7  & 89.7  & 37.6  & 42.0  & 76.1  & 69.7  & 77.6  & 55.2  & 47.3  & 56.7  \\
    QVQ-72B-Preview$\dagger$ & 87.3  & 71.1  & 66.2  & 76.7  & 45.8  & 67.9  & 82.5  & 42.1  & 47.4  & 74.0  & 71.2  & 65.3  & 58.5  & 52.9  & 48.2  \\
    \bottomrule
    \end{tabular}%
  \caption{Mathematical Evaluation of more MLLMs on Different Subjects in GeoSense-English. GPI = Geometric Principles Identification, GPA= Geometric Principles Application, Calculation of Solid Figures = CSF, Understanding of Solid Figures = USF, Transformation and Motion of Plane Figures = TMPF, Calculation of Plane Figures = CPF and Understanding of Plane Figures = UPF. MLLMs with $\dagger$ are typically trained for reasoning tasks.}
  \label{tab:en_all_results2}%
\end{table*}%
\end{document}